\documentclass[10pt,twocolumn,letterpaper]{article}

\usepackage{3dv}

\usepackage{tikz}
\usetikzlibrary{positioning}

\usepackage{times}
\usepackage{epsfig}
\usepackage{graphicx}
\usepackage{amsmath}
\usepackage{amssymb}
\usepackage{cuted}
\usepackage{capt-of}
\usepackage{subcaption}
\usepackage{multirow}
\usepackage{siunitx}
\usepackage{xcolor}
\usepackage{import}

\usepackage[pagebackref=true,breaklinks=true,letterpaper=true,colorlinks,bookmarks=false]{hyperref}

\newcommand{\papertitle}{DSP-SLAM: Object Oriented SLAM with Deep Shape Priors}

\newcommand\ignore[1]{}

\newcommand{\norm}[1]{\left\lVert#1\right\rVert}

\def \path{\bp C}

\newcommand{\matr}[1]{\mathbf{#1}}

\newcommand{\BOmega}{\mathbf{\Omega}}

\newcommand{\calB}{{\mathcal{B}}}

\newcommand{\calD}{{\mathcal{D}}}

\newcommand{\calM}{{\mathcal{M}}}

\threedvfinalcopy 

\def\threedvPaperID{347} 

\ifthreedvfinal\fi
\begin{document}

\title{\papertitle}

\author{Jingwen Wang \hspace{0.7cm}
Martin Rünz \hspace{0.7cm}
Lourdes Agapito \\
Department of Computer Science, University College London\\
{\tt\small \{jingwen.wang.17, martin.runz.15, l.agapito\}@ucl.ac.uk}
}

\twocolumn[{%
\renewcommand\twocolumn[1][]{#1}%
    \maketitle
    \vspace{-0.6cm}
    \centering
    \includegraphics[width=1.00\textwidth]{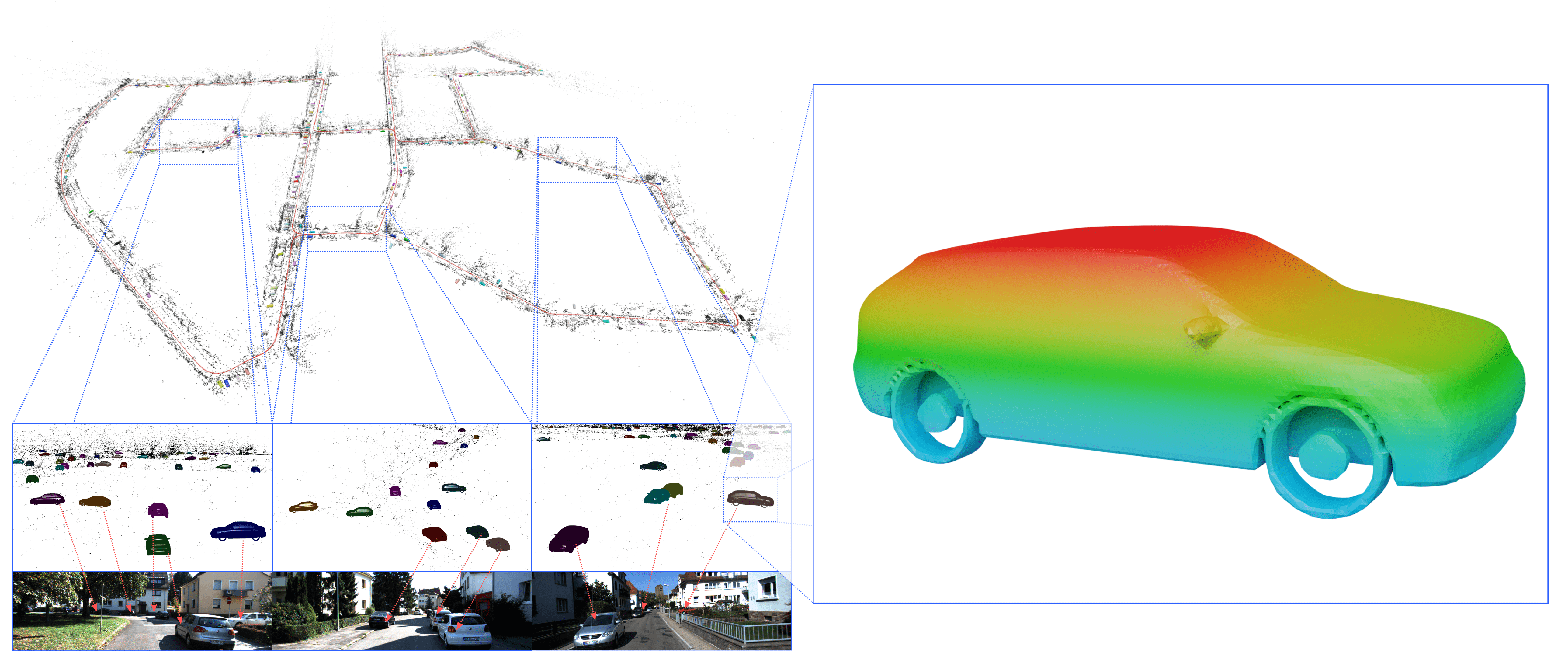}
    \captionof{figure}{DSP-SLAM builds a rich object-aware map, providing complete detailed shapes of detected objects, while representing the background coarsely as sparse feature points. Reconstructed map and camera trajectory on KITTI 00. }
    \label{figure-teaser}
    \vspace{0.5cm}
}]


\begin{abstract}
\vspace{-0.2cm}
We propose DSP-SLAM, an object-oriented SLAM system that builds a rich and accurate joint map of dense 3D models for foreground objects, and sparse landmark points to represent the background. DSP-SLAM takes as input the 3D point cloud reconstructed by a feature-based SLAM system and equips it with the ability to enhance its sparse map with dense reconstructions of detected objects. Objects are detected via semantic instance segmentation, and their shape and pose are estimated using category-specific deep shape embeddings as priors, via a novel second order optimization.  Our object-aware bundle adjustment builds a pose-graph to jointly optimize camera poses, object locations and feature points. DSP-SLAM can operate at $10$ frames per second on $3$ different input modalities: monocular, stereo, or stereo+LiDAR. We demonstrate DSP-SLAM operating at almost frame rate on monocular-RGB sequences from the Friburg and Redwood-OS datasets, and on stereo+LiDAR sequences on the KITTI odometry dataset showing that it achieves high-quality full object reconstructions, even from partial observations, while maintaining a consistent global map. Our  evaluation shows improvements in object pose and shape reconstruction with respect to recent deep prior-based reconstruction methods and reductions in camera tracking drift on the KITTI dataset. More details and demonstrations are available at our project page: \url{https://jingwenwang95.github.io/dsp-slam/}
\vspace{-0.5cm}
\end{abstract}

\begin{figure*}[t]
\centering
\includegraphics[width=1.00\textwidth]{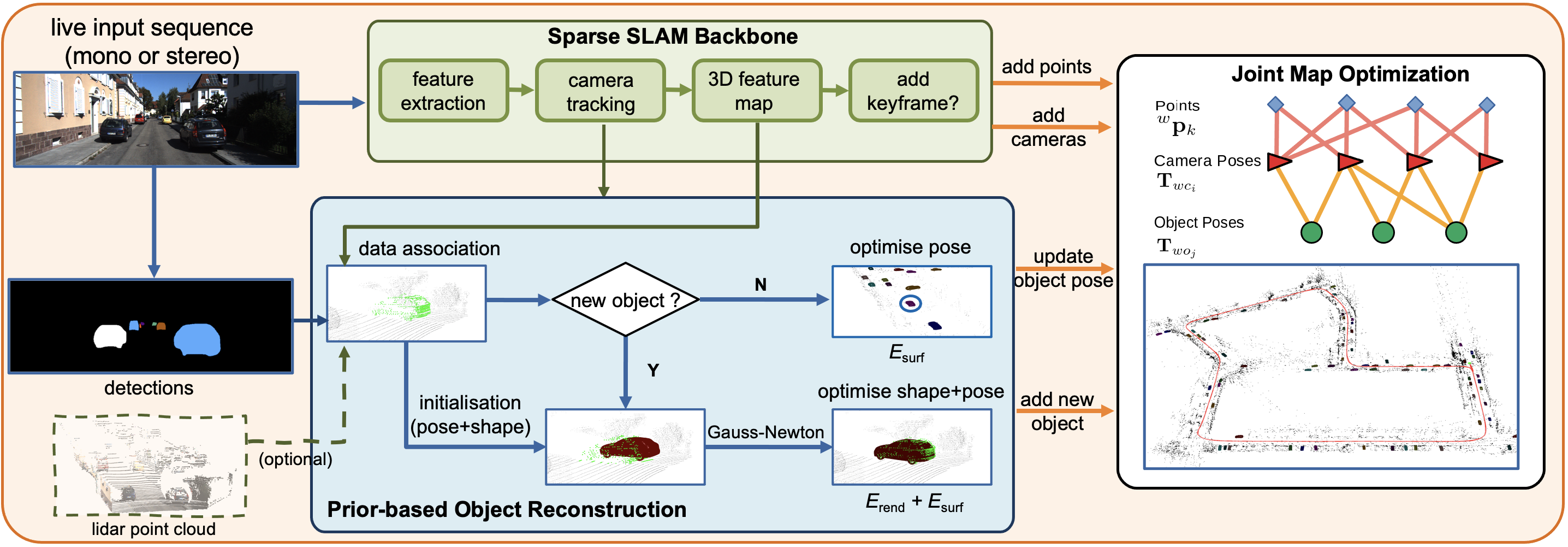}
\caption{System overview: DSP-SLAM takes a live stream of monocular or stereo images, infers object masks, and outputs a joint map of feature points and dense objects. The sparse SLAM backbone provides per-frame camera poses and a 3D point cloud.  At each  keyframe, a shape code is estimated for each new detected object instance, using a combination of 3D surface consistency and rendered depth losses. DSP-SLAM can operate in $3$ different modes: monocular, stereo, and stereo+LiDAR (when an optional LiDAR point cloud is available). \vspace{-0.2cm}
}
\label{fig:system}
\end{figure*}

\section{Introduction}
Simultaneous Localization and Mapping (SLAM) is the process of estimating the trajectory of a moving camera while reconstructing its surrounding environment. From a purely geometric perspective, SLAM is often regarded as a well-understood or even solved problem. Many state-of-the-art dense SLAM algorithms can achieve accurate trajectory estimation and create high-quality geometric reconstructions that can be used in obstacle avoidance or path planning for mobile robots. However, when it comes to more complex tasks that require scene understanding, geometry-only scene representations fall short of providing key semantic information. 
%
%
Taking advantage of recent deep learning breakthroughs in semantic segmentation and object detection algorithms \cite{maskrcnn, faster-rcnn, yolo} semantic SLAM systems augment geometric low-level map primitives by fusing semantic labels into the 3D reconstruction~\cite{semantic-fusion, semantic-stereo-fusion,Chen2019SuMa++}. However, the resulting scene maps merely consist of a set of labelled 3D points where reasoning about the scene at the level of objects to infer meaningful information such as the number of objects of each category, their size, shape or relative pose remains a challenging task. Better use of the semantic information is required in the form of an object-centric map representation that allows detailed shape estimation and meaningful instantiation of scene objects. 


Our proposed approach forms part of a more recent family of \emph{object-aware} SLAM methods that reconstruct object-centric maps grouping all the low-level geometric primitives (voxels, points ...) that make up the same object into a single instance. Front-end camera tracking and back-end optimization are both performed at the level of object instances. While the first object-level methods, such as SLAM++~\cite{SLAM++}, mapped previously known object instances, more recent systems have taken advantage of instance level semantic segmentation masks \cite{maskrcnn} to achieve object level reconstruction for unknown objects \cite{fusion++} even in the presence of dynamic objects  \cite{maskfusion, midfusion}.

\begin{figure}[tbp]
\centering
\includegraphics[width=0.235\textwidth]{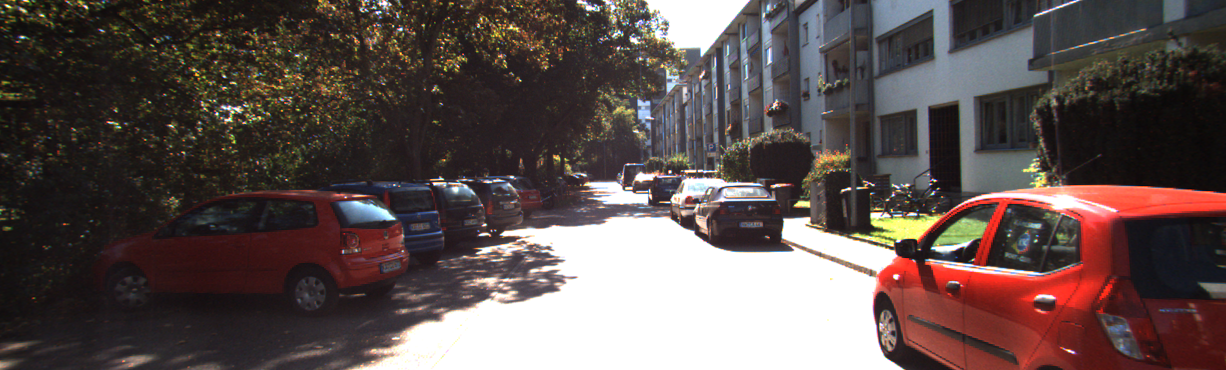}
\includegraphics[width=0.235\textwidth]{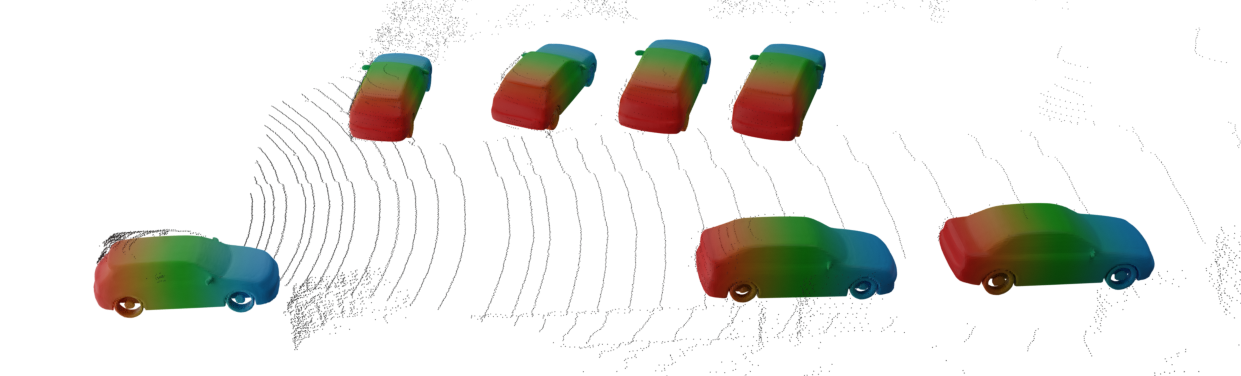}
\caption{Qualitative shape and pose results on a stereo+LiDAR KITTI sequence. A very sparse set of LiDAR points was used to reconstruct each car. LiDAR points on the road are only shown for illustration.\vspace{-0.5cm}}
\label{fig:single-view}
\end{figure}

However, these early object level SLAM systems exhibit major drawbacks: They either require a pre-known database of object instances \cite{SLAM++}; or reconstruct objects from scratch without exploiting shape priors \cite{maskfusion, midfusion, fusion++}, which results in partial or incomplete object reconstructions. We improve this by exploiting the regularity of shapes within an object category in the form of learned shape priors, defined as a latent code $\matr{z}$ and a generative model $G(\matr{z})$ that decodes it into its full geometry. This brings us several advantages; object shapes decoded from latent codes are guaranteed to be detailed and complete, regardless of partial observations or changes in view-points, they provide a compact representation and they can be optimized using the gradients obtained through back-propagation. 

Using ORB-SLAM2~\cite{orbslam2} as a sparse camera tracking and mapping backbone, DSP-SLAM takes the reconstructed 3D point-cloud as input and fits a latent code to each detected object instance, using a combination of 3D surface consistency and rendered depth losses. Foreground objects, background features and camera poses are further refined via bundle adjustment using a joint factor graph. We show DSP-SLAM  operating in $3$ different modes: monocular, stereo, and stereo+LiDAR. The monocular and stereo systems use the respective ORB-SLAM2 modalities as the SLAM backbone and the reconstructed 3D point-clouds to reconstruct the detected objects. The stereo+LiDAR system uses stereo ORB-SLAM2 as the SLAM backbone but in addition it incorporates a sparse set of LiDAR measurements (as few as 50 per object) for object reconstruction and pose-only optimization. 

\noindent\textbf{Contributions:} While DSP-SLAM is not the first approach to leverage shape priors for 3D reconstruction~\cite{frodo,node-slam} from image sequences, it innovates in various ways. 
%
%
Firstly, unlike~\cite{frodo,node-slam}, our map does not only represent objects, but also reconstructs the background as sparse feature points, optimizing them together in a joint factor graph, marrying the best properties of feature-based~\cite{orbslam2} (highly accurate camera tracking) and object-aware SLAM (high level semantic map). Secondly, although Node-SLAM \cite{node-slam} also incorporates shape priors within a real-time SLAM system~\cite{node-slam}, it uses dense depth images for shape optimization, while DSP-SLAM can operate with RGB-only monocular streams and requires as few as $50$ 3D points per object to obtain accurate shape estimates. Finally, although both FroDO~\cite{frodo} and DSP-SLAM can operate in a monocular RGB setting, FroDO is a slow batch approach that requires all frames to be acquired in advance and associated with their camera poses, while  
DSP-SLAM is an online, sequential method that can operate at $10$ frames per second. 

In terms of object shape and pose estimation, we improve quantitative and qualitatively over auto-labelling~\cite{Zakharov2020Autolabeling3D}, a state-of-the-art prior-based object reconstruction method. Experiments on the KITTI odometry \cite{kitti_odom} dataset show that, with stereo+LiDAR input our joint bundle adjustment offers improvements in trajectory estimation over the feature-only stereo system ORB-SLAM2~\cite{orbslam2}, used as our backbone. Moreover, DSP-SLAM offers comparable tracking performance to state-of-the-art stereo~\cite{Wang2017StereoDSO}, LiDAR-only~\cite{Chen2019SuMa++} and dynamic~\cite{Bescos2018DynaSLAM} SLAM systems, while providing rich dense object reconstructions. DSP-SLAM also achieves promising qualitative reconstruction results with monocular input on Freiburg Cars \cite{freiburg-cars} and Redwood-OS \cite{choi2016redwood} dataset.


\section{Related work} 

\noindent\textbf{Object-aware SLAM:}
SLAM++ \cite{SLAM++} pioneered object-aware RGB-D SLAM, representing the scene at the level of objects using a joint pose-graph for camera and object poses. A database of pre-scanned objects was created in advance and object instances were detected and mapped using a pre-trained 3D detector, ICP losses and pose-graph optimization. In later work, Tateno \emph{et al.}~\cite{Tateno2016When2I} aligned object instances from a pre-trained database to volumetric maps while Stuckler \emph{et al.}~\cite{stuckler2012model} performed online exploration, learning object models on the fly and tracking them in real time. An important drawback of instance-based approaches is their inability to scale to a large number of objects and their need for object models to be known in advance. More recent object-aware RGB-D SLAM systems have dropped the requirement for known models and instead take advantage of state-of-the art 2D instance-level semantic segmentation masks~\cite{maskrcnn} to obtain object-level scene graphs~\cite{semantic-fusion} and per-object reconstructions via depth fusion, even in the case of dynamic scenes~\cite{maskfusion,midfusion}. 

Extensions of object-aware SLAM to the case of monocular video input deal with the additional challenge of relying only on RGB-only information~\cite{mono-object-slam, mono-object-sparse-slam, quadricslam, cubeslam, category-specific-models} which results in the use of simplistic shape representations. In QuadricSLAM \cite{quadricslam} objects are represented as ellipsoids and fit to monocular observations while in CubeSLAM \cite{cubeslam} cuboid proposals generated from single-view detections are optimized in a joint bundle adjustment optimization. 

While the above SLAM systems represent an important step forward towards equipping robots with the capability of building semantically meaningful object-oriented maps, they fall short of exploiting semantic priors for object reconstruction. In this paper we take this direction of using a category-specific learnt shape prior and embed this within an object-aware SLAM system. 

\noindent\textbf{3D Reconstruction with Shape Priors:}
The use of learnt compact shape embeddings as priors for 3D reconstruction has a long tradition in computer vision. From 3D morphable models for the reconstruction of faces or bodies~\cite{blanz1999morphable,loper2015smpl}, to PCA models to represent category specific object shape priors~\cite{directshape}. 
%
%
Other examples of the use of embedding spaces for single or multi-view shape reconstruction include GPLVMs \cite{dame2013dense, prisacariu2011shared, prisacariu2012simultaneous} or neural representations~\cite{hu2018dense-embedding, zhu2018object} such as a variational autoencoder~\cite{node-slam},  AltlasNet~\cite{AtlasNet,2019-lin} or DeepSDF~\cite{deepsdf,frodo,2019-xu,Zakharov2020Autolabeling3D}. DeepSDF \cite{deepsdf} provides a powerful implicit learnt shape model that encapsulates the variations in shape across an object category, in the form of an auto-decoder network that regresses the signed distance function (SDF) values of a given object and has been used as a shape prior for single-view~\cite{2019-xu} and multi-view~\cite{frodo} reconstruction. Similarly to~\cite{Zakharov2020Autolabeling3D} DSP-SLAM adopts DeepSDF as the shape prior and takes sparse LiDAR and images as input, however~\cite{Zakharov2020Autolabeling3D} takes single frames and is not a SLAM method. DOPS~\cite{Najibi2020DOPSLT} is a single-pass 3D object detection architecture for LiDAR that estimates both 3D bounding boxes and shape.  

Our approach is most closely related to those that build consistent multi-object maps over an entire sequence such as FroDO~\cite{frodo} and Node-SLAM~\cite{node-slam}. Unlike FroDO ~\cite{frodo} ours is a sequential SLAM system and not a batch method. Unlike Node-SLAM \cite{node-slam}, in our system low-level point features and high-level objects are jointly optimized to bring the best of both worlds: accurate tracking and rich semantic shape information. DeepSLAM++~\cite{hu2019deep-slam++} leverages shape priors in a SLAM pipeline by selecting 3D shapes predicted by Pix3D~\cite{Sun_2018_CVPR}, but forward shape generation is often unstable and lead to poor results on real data. 
\section{\label{system-overview}System Overview}

DSP-SLAM is a sequential localisation and mapping method that reconstructs the complete detailed shape of detected objects while representing the background coarsely as a sparse set of feature points. Each object is represented as a compact and optimizable code vector $\matr{z}$. We employ DeepSDF \cite{deepsdf} as the shape embedding, that takes as input a shape code $\matr{z} \in \mathbb{R}^{64}$ and a 3D query location $\matr{x} \in \mathbb{R}^{3}$, and outputs the signed distance function (SDF) value $s = G(\matr{x}, \matr{z})$ at the given point. An overview of DSP-SLAM is shown in Fig.~\ref{fig:system}. DSP-SLAM runs at almost real time ($10$ frames per second) and can operate on different modalities: monocular, stereo or stereo with LiDAR; depending on the available input data. 

\begin{figure}[tbp]
\centering
\includegraphics[width=0.115\textwidth]{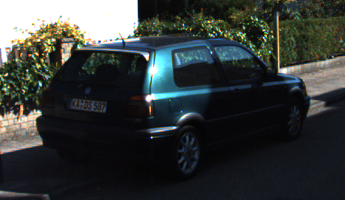} \hspace{-0.6em}
\includegraphics[width=0.115\textwidth]{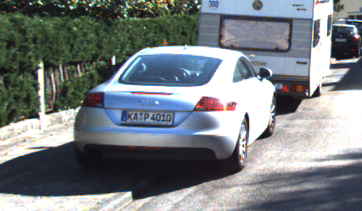} \hspace{-0.6em}
\includegraphics[width=0.115\textwidth]{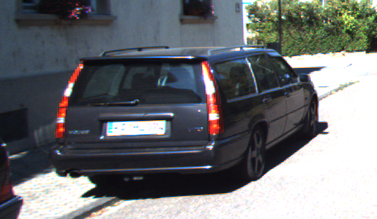} \hspace{-0.6em}
\includegraphics[width=0.115\textwidth]{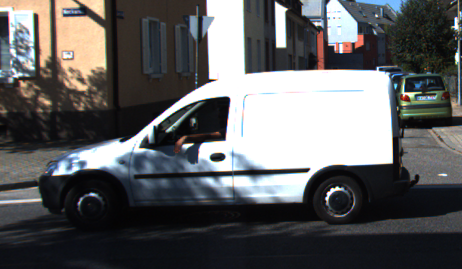} \hspace{-0.6em}
\includegraphics[width=0.115\textwidth]{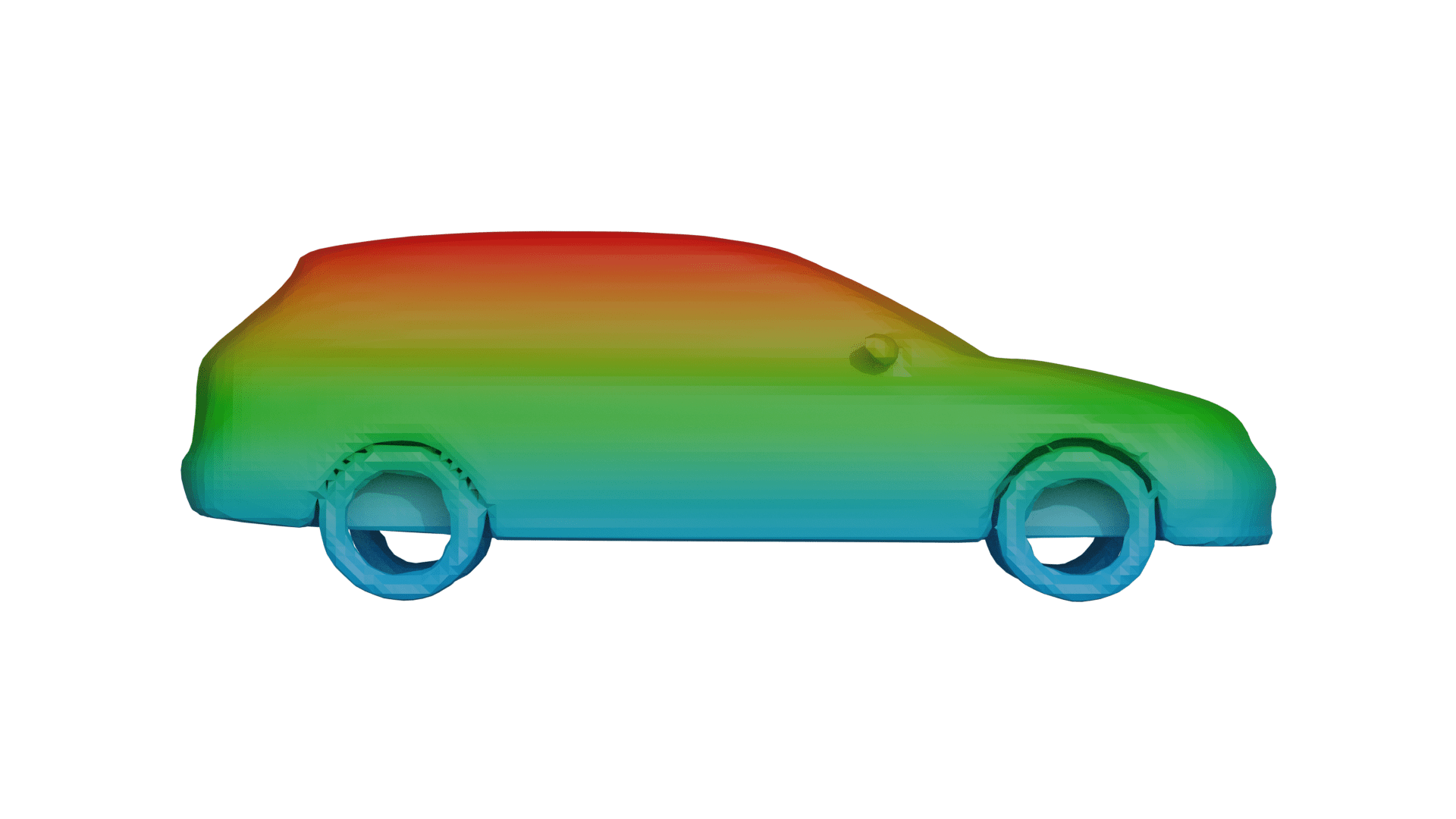} \hspace{-0.6em}
\includegraphics[width=0.1115\textwidth]{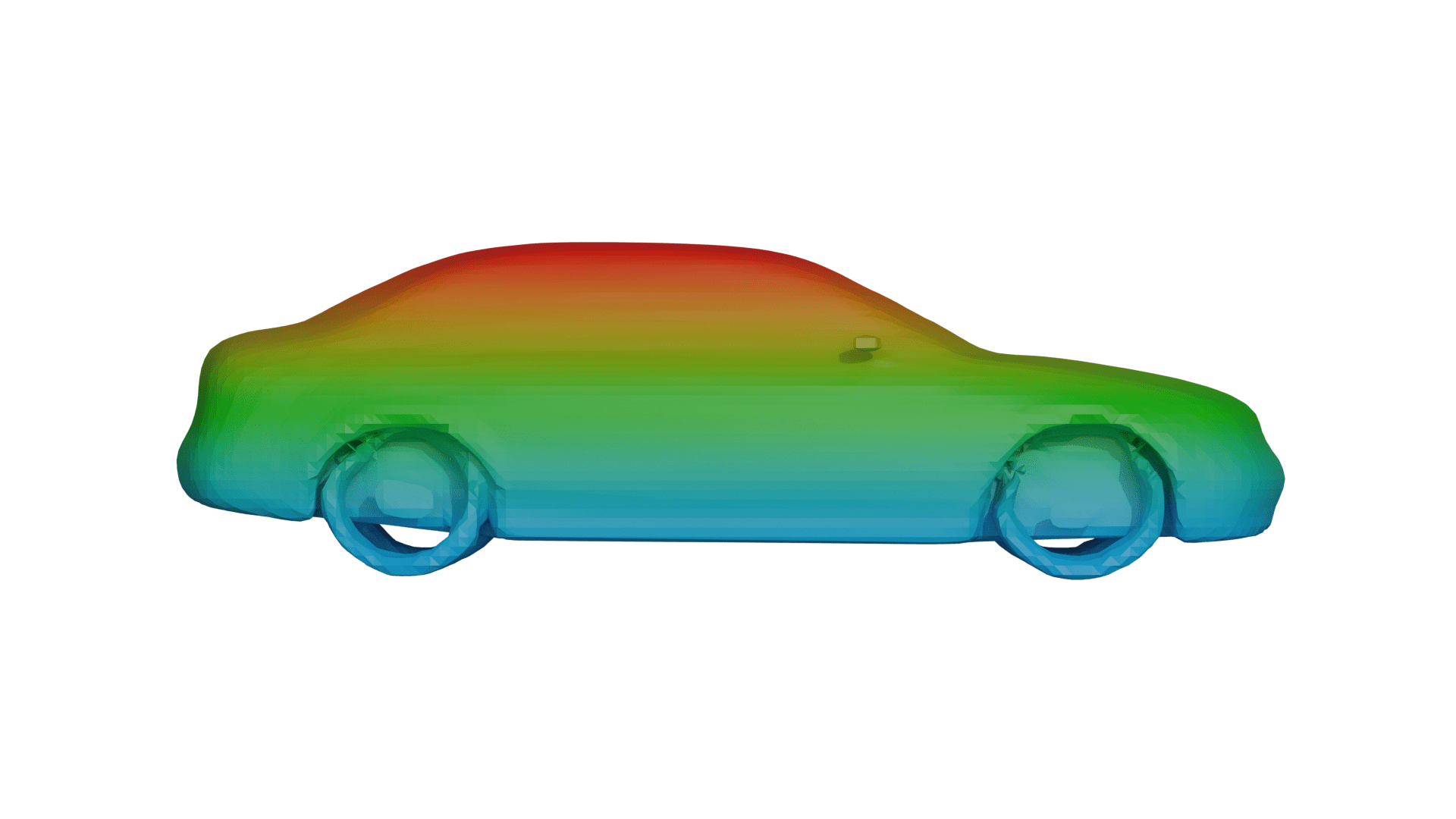} \hspace{-0.6em}
\includegraphics[width=0.115\textwidth]{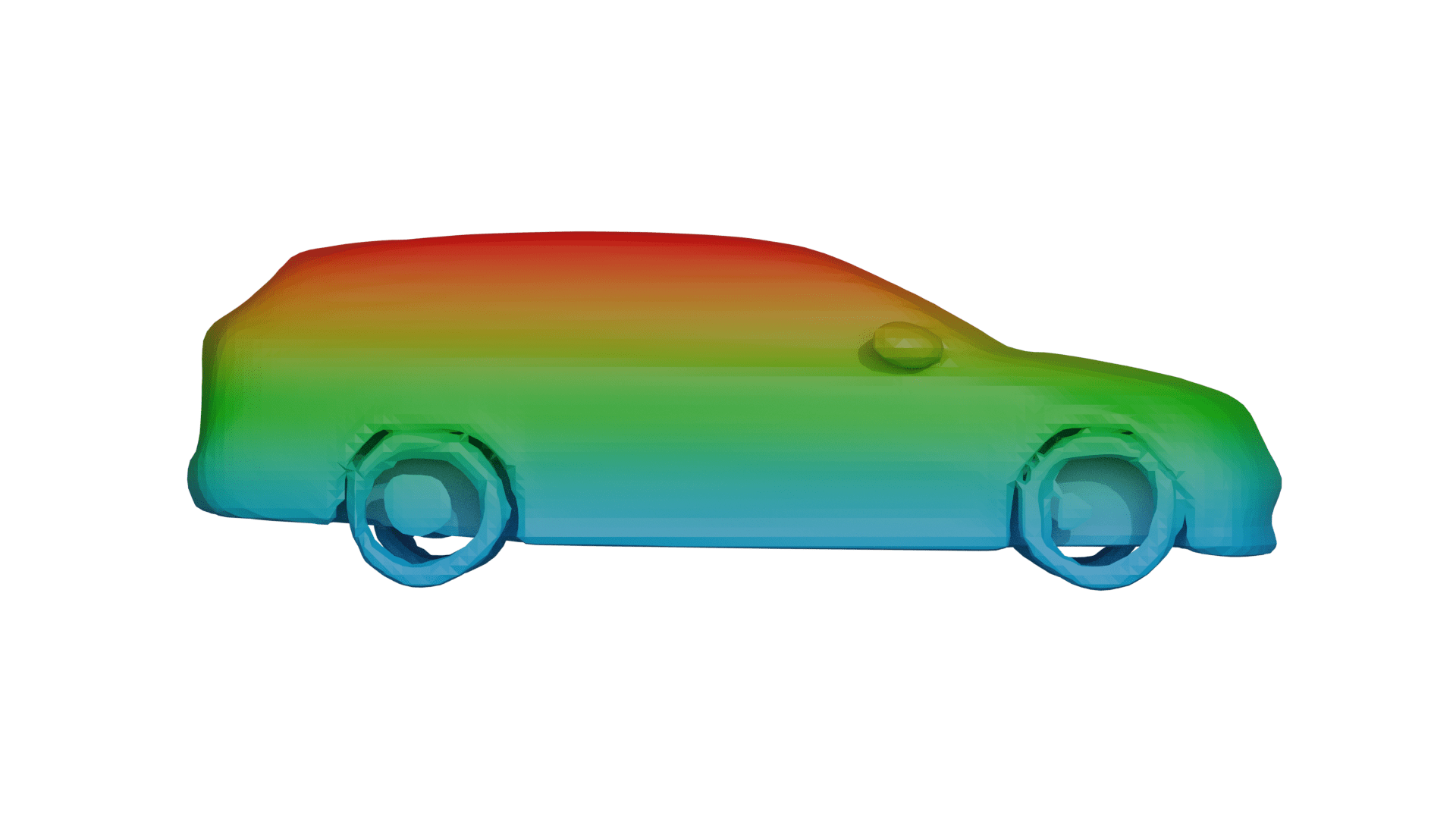} \hspace{-0.6em}
\includegraphics[width=0.115\textwidth]{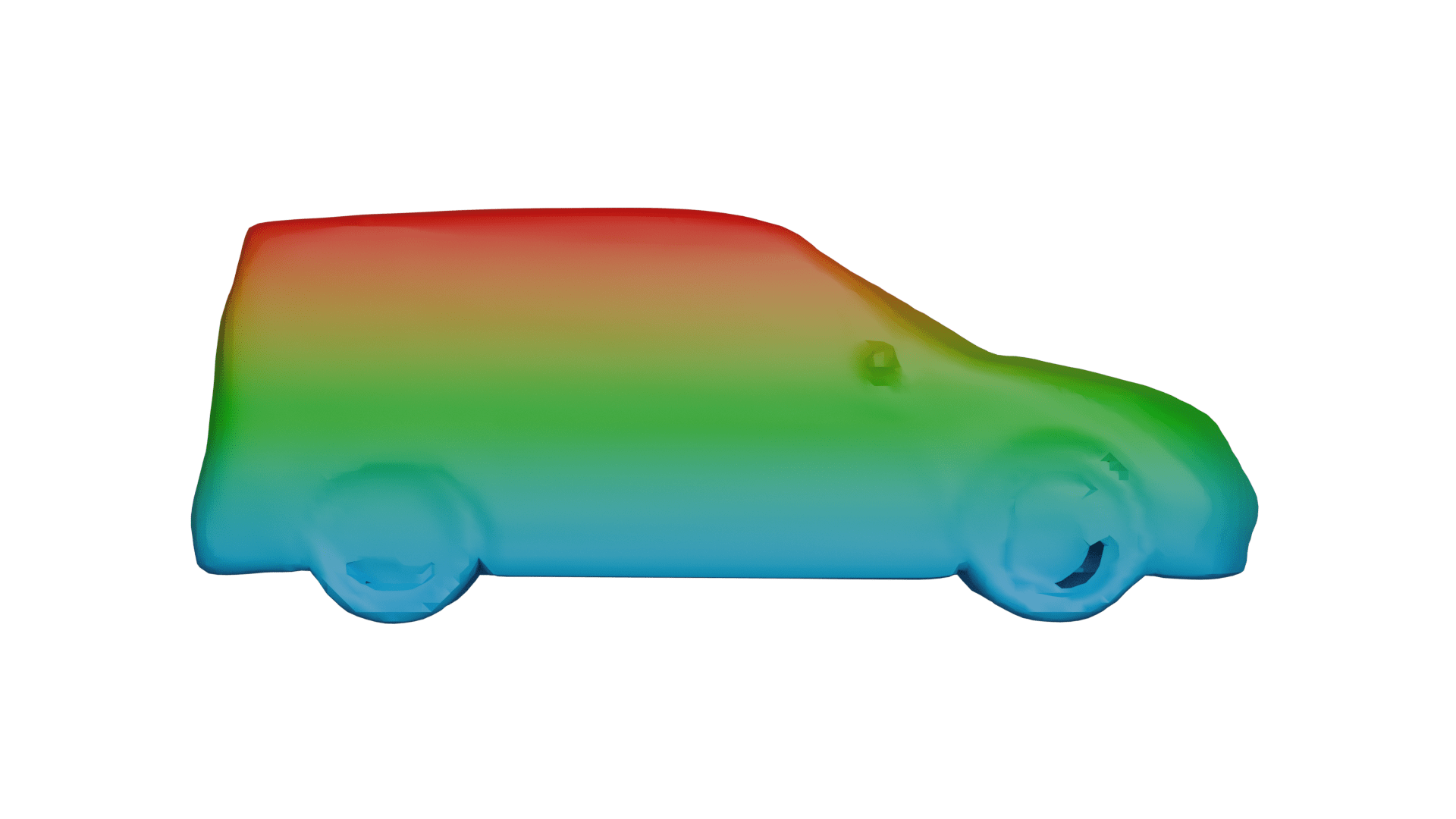} \hspace{-0.6em}
\caption{Shape reconstruction: qualitative results. \vspace{-0.5cm}}
\label{fig:single-view-shape}
\end{figure}

\noindent\textbf{Sparse SLAM backbone:} ORB-SLAM2~\cite{orbslam2} is used as the tracking and mapping backbone, a feature-based SLAM framework that can operate on monocular or stereo sequences. While the tracking thread estimates camera pose at frame-rate from correspondences, the mapping thread builds a sparse map by reconstructing 3D landmarks.

\noindent\textbf{Detections:}  We perform object detection at each key-frame, to jointly infer 2D bounding boxes and segmentation masks. In addition, an initial estimate for the object pose estimation is obtained via 3D bounding box detection~\cite{smoke,second}.

\noindent\textbf{Data association:} New detections will either be associated to existing map objects, or instantiated as a \textit{new} object via object-level data association. 
Each detected object instance $I$ consists of a 2D bounding box $\calB$, a 2D mask $\calM$, the dpeth observation of sparse 3D point cloud $\calD$, and the initial object pose $\matr{T}_{co, 0}$. 

\noindent\textbf{Prior-based object reconstruction:} Newly instantiated objects will be reconstructed following the object reconstruction pipeline described in Sec.~\ref{object-recon}. DSP-SLAM takes the set of sparse 3D point observations $\calD$, which can come from reconstructed SLAM points (in monocular and stereo modes) or LiDAR input (in stereo+LiDAR mode) and optimises the shape code and object pose to minimise surface consistency and depth rendering losses. Objects already present in the map will only have their 6-dof pose updated via pose-only optimization.

\noindent\textbf{Joint map optimisation:} A joint factor graph of point features (from SLAM), objects and camera poses is optimised via bundle adjustment to maintain a consistent map and incorporate loop closure. New objects are added as nodes to the joint factor graph and their relative pose estimates $\matr{T}_{co}$ as camera-object edges. Object-level data association and joint bundle adjustment will be discussed in Sec.~\ref{object-slam}.




\section{\label{object-recon}Object Reconstruction with  Shape Priors}

We aim to estimate the full dense shape $\matr{z}$ and 7-DoF  pose $\matr{T}_{co}$, represented as a homogeneous transformation matrix $\matr{T}_{co} = [s\matr{R}_{co}, \matr{t}_{co}; \matr{0}, 1] \in \matr{Sim}(3)$, for an object with detections $I = \{\calB, \calM, \calD, \matr{T}_{co, 0}\}$. We formulate this as a joint optimization problem, which iteratively refines the shape code and object pose from an initial estimate. We propose two energy terms $E_{surf}$ and $E_{rend}$ and formulate a Gauss-Newton solver with analytical Jacobians.

\subsection{Surface Consistency Term}
\begin{figure}[tbp]
    \centering
    \includegraphics[width=0.48\textwidth]{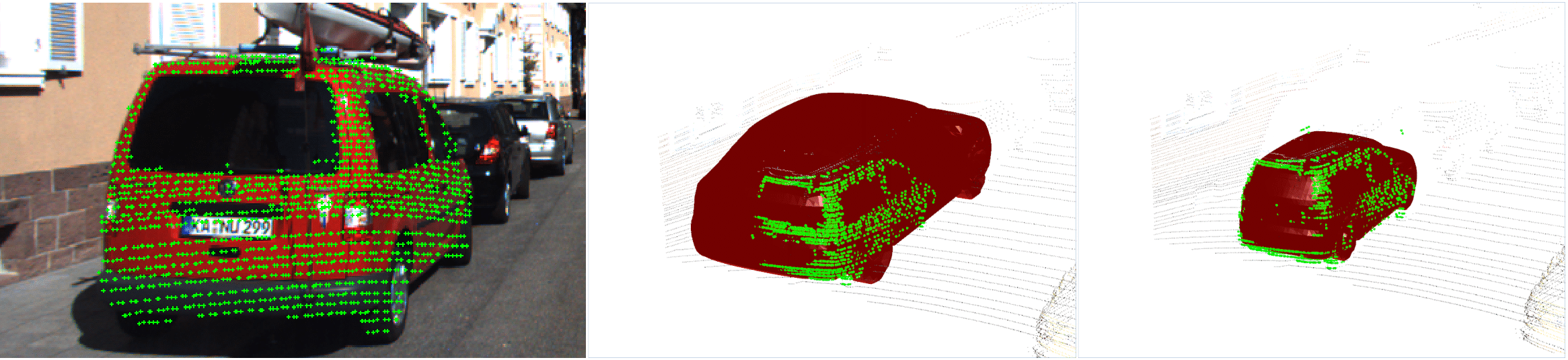}
    \caption{Illustration of the effectiveness of the rendering term in the presence of partial observations. \textbf{Left:} Detected object and partial surface point observations (green). \textbf{Middle:} Optimisation result with $E_{surf}$ only. The loss is minimised but the shape grows larger than its actual size. \textbf{Right:} Optimisation result with the rendering term. Enforcing the silhouette constraint results in the correct scale. \vspace{-0.6cm}}
    \label{fig:shape-ablation}
\end{figure}

This term measures the alignment between observed 3D points and the reconstructed object surface. 
\begin{equation} \label{E_surf}
    E_{surf} = \frac{1}{\left| \BOmega_s \right|}\sum_{\matr{u} \in \BOmega_s} G^2(\matr{T}_{oc} \pi^{-1}(\matr{u}, \calD), \matr{z})
\end{equation}
where $\BOmega_s$ denotes the pixel coordinates of the set of sparse 3D points $\calD$, which can come from reconstructed SLAM points (in monocular and stereo modes) or LiDAR input (in stereo+LiDAR mode). Ideally, the back-projected point at pixel $\matr{u}$ should perfectly align with the object surface resulting in zero SDF value, giving a zero error residual. In practice, we observed that the surface consistency term alone is not sufficient for correct shape and pose estimation in the case of partial observations. Fig.~\ref{fig:shape-ablation} illustrates a case where only points on the back and the right side of a car are observed (shown in green). Using  the surface consistency alone term leads to incorrect shape estimation -- much larger than its actual size. To address this issue, we propose a rendering loss, that provides point-to-point depth supervision and enforces silhouette consistency to penalize shapes that grow outside of the segmentation mask. 

\subsection{Differentiable SDF Renderer}

Following \cite{multi-view_supervision, node-slam}, we build our SDF renderer via differentiable ray-tracing. For each pixel $\matr{u}$, we back-project a ray $^c\matr{x} = \matr{o} + d \matr{K}^{-1} \dot{\matr{u}}$ parameterized by the depth value $d$ under camera coordinate space, with $\matr{o}$ being the camera optical centre and $\matr{K}$ being camera intrinsic matrix. We sample $M$ discrete depth values $\{d_i\}$ along each ray within the range $[d_{min}, d_{max}]$, with $d_i = d_{min} + (i-1) \Delta d$, and $\Delta d = (d_{max} - d_{min}) / (M-1)$. The bounds of the depth range are determined by the current estimation of object translation and scale, and are re-computed at each iteration.

\noindent\textbf{Occupancy Probabilities} 
The SDF value $s_i$ at each sampled point can be obtained by transforming sampled points to the object coordinate frame and passing through the DeepSDF decoder. The SDF value encodes the probability that a given point is occupied by the object or belongs to free space. We apply a piecewise linear function to the predicted SDF values to indicate the occupancy probability $o_i$, defined in Eq.~\ref{sdf2occ}, where $\sigma$ represents the cut-off threshold which controls the smoothness of the transition. We fix $\sigma = 0.01$ throughout our experiments. \vspace{-0.4cm}

\begin{equation} \label{sdf2occ}
    s_i = G(\matr{T}_{oc} {^c\matr{x}},\, \matr{z}) \quad \textrm{and} \quad 
    o_i = 
    \begin{cases}
        1 & s_i < -\sigma\\
        -\frac{s_i}{2 \sigma} &  \left|s_i\right| \leq \sigma \\    
        0 & s_i > \sigma
    \end{cases}
\end{equation}

\noindent\textbf{Event Probabilities} When tracing points along the ray, the ray either terminates or escapes without hitting other points. These $M+1$ event probabilities can be defined as: 
\begin{eqnarray} \label{term_prob}
    & \phi_i = o_i \prod_{j=1}^{i-1} (1- o_j), i = 1, \dots, M \nonumber \\
    & \phi_{M + 1} = \prod_{j=1}^{M} (1- o_j)
\end{eqnarray}

\noindent\textbf{Rendered Depth and Rendering Term} With the probabilities defined above, the rendered depth value at each pixel $\matr{u}$ can be computed as the expected depth value of the terminating point as in Eq.~\ref{render_depth}. To make it consistent, we set $d_{M+1}$, the depth value associated with escape probability, to a constant value $1.1 d_{max}$, as in \cite{node-slam}.
\begin{equation} \label{render_depth}
    \hat{d}_{\matr{u}} = \sum_{i=1}^{M+1} \phi_i d_i
\end{equation}
Since the rendering is fully differentiable, it can be integrated in our optimization. Unlike~\cite{node-slam, multi-view_supervision}, we perform  ray-tracing in continuous space and do not require to discretize the object model. The final rendering term is as follows:
\begin{equation}
    E_{rend} = \frac{1}{\left| \BOmega_r \right|} \sum_{\matr{u} \in \BOmega_r} (d_\matr{u} - \hat{d}_{\matr{u}})^2
\end{equation}
where $\BOmega_r = \BOmega_s \cup \BOmega_b$ is the union of surface pixels and pixels not on object surface but inside the 2D bounding box $\calB$. Surface pixels $\BOmega_s$ are the same set of pixels used in Eq.~\ref{E_surf}, obtained by projecting the 3D reconstucted SLAM points onto the image masks as discussed in Sec.~\ref{system-overview}. The pixels in $\BOmega_b$ are assigned the same depth value as $d_{M+1} = 1.1 d_{max}$ and provide important silhouette supervision for our optimization since they penalize renderings that lie outside the object boundary, forcing empty space. As the pixels in $\BOmega_b$ do not require a depth measurement, we perform uniform sampling inside the 2D bounding box and filter out those inside the segmentation mask.

\subsection{Optimization details}

Our final energy is the weighted sum of the surface and rendering terms and a shape code regularization term:
\begin{equation}
    E = \lambda_s E_{surf} + \lambda_r E_{rend} + \lambda_c \norm{\matr{z}}^2
\end{equation}
The hyperparameter values used for optimization $\lambda_s = 100$, $\lambda_r = 2.5$ and $\lambda_c = 0.25$ were tuned such that the Hessian matrices of the energy terms are of the same order of magnitude. Since all terms are quadratic, we adopt a  Gauss-Newton optimisation approach with analytical Jacobians (Please refer to supplemental material for detail), initialized from a zero shape code $\matr{z} = \matr{0}$. The initialisation for the object pose 
$\matr{T}_{co, 0}$  is given by a LiDAR 3D detector~\cite{second} when LiDAR is available. In the monocular/stereo case, it is given by an image-based 3D detector~\cite{smoke} or by performing PCA on the sparse object point cloud. 

\section{\label{object-slam}Object SLAM}

As an object-based SLAM system, DSP-SLAM builds a joint factor graph of camera poses, 3D feature points and object locations and poses. As Fig.~\ref{system-overview} shows, the factor graph introduces object nodes and camera-object edges. 


\subsection{\label{data-association}Object Data Association}

Data association between new detections and reconstructed objects is an important step in object-level SLAM. We aim to associate each detection $I$ to its \textit{nearest} object $o$ in the map, adopting different strategies depending on the different input modalities. When LiDAR input is available we compare the distance between 3D bounding box and reconstructed object. When only stereo or monocular images are used as input, we count the number of matched feature points between the detection and object. If multiple detections are associated with the same object, we  keep the \textit{nearest} one and reject others. Detections not associated with any existing objects are initialised as new objects and their shape and pose  optimised following Sec.~\ref{object-recon}. For stereo and monocular input modes, reconstruction only happens when enough surface points are observed. For detections associated with existing objects, only the pose is optimised by running pose-only optimization and a new camera-object edge added to the joint factor-graph.

\begin{table*}
\renewcommand{\arraystretch}{1.0}
\centering
\begin{tabular}{c|c c c c|c c c c}
\hline
\multirow{2}{*}{Diff.} & \multicolumn{4}{c|}{Auto-labelling\cite{Zakharov2020Autolabeling3D}} & \multicolumn{4}{c}{Ours} \\
\cline{2-9}
 & BEV@0.5 & 3D@0.5 & NS@0.5 & NS@1.0 & BEV@0.5 & 3D@0.5 & NS@0.5 & NS@1.0 \\
\hline
E & 80.70 & \textbf{63.96} & 86.52 & 94.31 & \textbf{83.31} & 62.58 & \textbf{88.01} & \textbf{96.86} \\ 
M & 63.36 & 44.79 & 64.44 & 85.24 & \textbf{75.28} & \textbf{47.76} & \textbf{76.15} & \textbf{89.97} \\
\hline
\end{tabular}
\caption{Quantitative comparison of object cuboid prediction quality with Auto-labelling on KITTI3D on Easy and Moderate samples. Results of Auto-labelling are taken from their paper. Best results are shown as bold numbers.
\vspace{-0.1cm}}
\label{tab:object_detection}
\end{table*}


\begin{figure*}[tbp]
\centering
\includegraphics[width=1.00\textwidth]{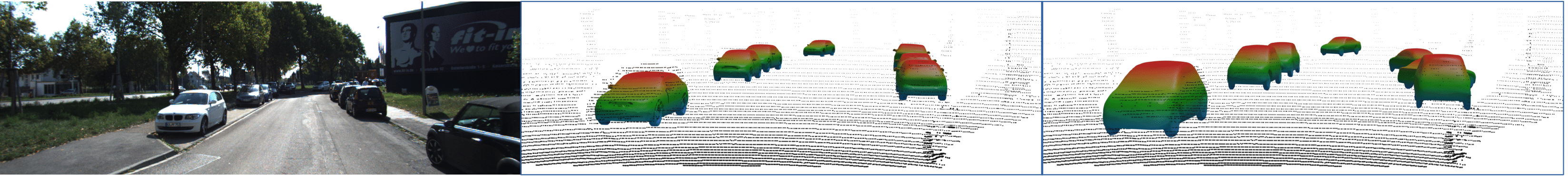}
\caption{A qualitative comparison of shape reconstruction and pose estimation against Auto-labelling~\cite{Zakharov2020Autolabeling3D}. \textbf{Left:} input RGB image. \textbf{Middle:} result with DSP-SLAM \textbf{Right:} result with auto-labelling~\cite{Zakharov2020Autolabeling3D}\vspace{-0.5cm}}
\label{fig:comparison-auto-labelling}
\end{figure*}

\subsection{\label{joint_ba}Joint Bundle Adjustment}

Our joint map consists of a set of camera poses $C = \{\matr{T}_{wc_i}\}_{i=1}^{M}$, object poses $O = \{\matr{T}_{wo_j}\}_{j=1}^{N}$ and map points $P = \{^w\matr{p}_k\}_{k=1}^{K}$. Our joint BA can be formulated as a non-linear least squares optimization problem:
\begin{eqnarray}
    C^*, O^*, P^* &{} = {}& \mathop{\arg\min}_{\{C, O, P\}} \sum_{i, j} \norm{\matr{e}_{co}(\matr{T}_{wc_i}, \matr{T}_{wo_j})}_{\Sigma_{i,j}} \nonumber \\
    && {+}\:\sum_{i, k} \norm{\matr{e}_{cp}(\matr{T}_{wc_i}, ^w\matr{p}_k)}_{\Sigma_{i,k}}
\label{eq:ba}
\end{eqnarray}
where $\matr{e}_{co}$ and $\matr{e}_{cp}$ represent the residuals for camera-object and camera-point measurements and $\Sigma$ is the co-variance matrix of measurement residuals. Objects act as additional landmarks, which results in improvements in tracking performance as shown in our evaluation on KITTI. The optimization is solved with Levenberg-Marquardt in g2o \cite{g2o}.

\noindent\textbf{Camera-Object Measurements:} Object-camera poses $\matr{T}_{co}$ are evaluated by minimising the surface alignment term in Eq. \ref{E_surf} while keeping  the shape code and scale fixed. New pose observations serve as edges between camera pose $\matr{T}_{wc}$ and object pose $\matr{T}_{wo}$, and the residual is defined as:
 $   \matr{e}_{co} = \log (\matr{T}^{-1}_{co} \cdot \matr{T}^{-1}_{wc} \cdot \matr{T}_{wo})$
%
where $\log$ is the logarithm mapping from $\matr{SE}(3)$ to $\mathfrak{se}(3)$. Poses in the factor graph are 6-DoF, as object scale is only optimised when first detected.

\noindent\textbf{Camera-Point Measurements:} We use the standard formulation of re-projection error used in ORB-SLAM2~\cite{orbslam2}:
$    \matr{e}_{cp} = \pi(\matr{T}_{wc}^{-1} {^w\matr{p}}) - \matr{\Tilde{u}}$, 
%
where $\matr{\Tilde{u}}$ is the measured pixel coordinate of map point $\matr{p}$. We follow a similar strategy as ORB-SLAM2 to tune $\Sigma_{ij}$ such that the two energy terms contribute roughly the same to the overall optimization. 
\section{Experimental Results}

\begin{table*}
\centering
\setlength{\arrayrulewidth}{0.5mm}
\resizebox{1.00\textwidth}{!}{
\begin{tabular}{c c c c c c c c c c c c c}
\hline
\multirow{3}{*}{Approach} & \multicolumn{11}{c}{Sequence} & \multirow{3}{*}{Average} \\
& 00* & 01 & 02* & 03 & 04 & 05* & 06* & 07* & 08* & 09* & 10 & \\
& rpe/rre & rpe/rre & rpe/rre & rpe/rre & rpe/rre & rpe/rre & rpe/rre & rpe/rre & rpe/rre & rpe/rre & rpe/rre & \\
\hline
SuMa++ \cite{Chen2019SuMa++} & \textbf{0.64}/\textbf{0.22} & 1.60/0.46 & 1.00/0.37 & 0.67/0.46 & \textbf{0.37}/0.26 & 0.40/0.20 & 0.46/0.21 & \textbf{0.34}/\textbf{0.19} & 1.10/0.35 & \textbf{0.47}/\textbf{0.23} & \textbf{0.66}/0.28 & \textbf{0.70}/0.29 \\
Ours St+LiDAR (250pts) & 0.75/\textbf{0.22} & \textbf{1.49}/0.20 & \textbf{0.79}/\textbf{0.23} & \textbf{0.60}/\textbf{0.18} & 0.47/0.11 & \textbf{0.32}/\textbf{0.15} & 0.39/0.21 & 0.52/0.28 & \textbf{0.94}/\textbf{0.27} & 0.79/0.28 & 0.69/\textbf{0.26} & \textbf{0.70}/\textbf{0.22} \\
Ours St+LiDAR (50pts) & 0.80/0.24 & 1.50/\textbf{0.15} & 0.84/0.26 & 0.61/0.18 & 0.44/\textbf{0.10} & 0.32/0.16 & \textbf{0.35}/\textbf{0.15} & 0.57/0.24 & 1.03/0.30 & 0.78/0.27 & 0.67/0.30 & 0.72/\textbf{0.22} \\
\hline
ORB-SLAM2 \cite{orbslam2} & 0.70/0.25 & \textbf{1.38}/0.20 & 0.76/0.23 & 0.71/0.17 & 0.45/0.18 & \textbf{0.40}/\textbf{0.16} & 0.51/0.15 & \textbf{0.50}/\textbf{0.28} & 1.07/0.31 & \textbf{0.82}/0.25 & 0.58/0.28 & \textbf{0.72}/0.22 \\
St DSO \cite{Wang2017StereoDSO} & 0.84/0.26 & 1.43/\textbf{0.09} & 0.78/\textbf{0.21} & 0.92/\textbf{0.16} & 0.65/0.15 & 0.68/0.19 & 0.67/0.20 & 0.83/0.36 & \textbf{0.98}/\textbf{0.25} & 0.98/\textbf{0.18} & \textbf{0.49}/\textbf{0.18} & 0.84/\textbf{0.20} \\
St LSD-SLAM \cite{engel2015stereo-lsd} & \textbf{0.63}/0.26 & 2.36/0.36 & 0.79/0.23 & 1.01/0.28 & \textbf{0.38}/0.31 & 0.64/0.18 & 0.71/0.18 & 0.56/0.29 & 1.11/0.31 & 1.14/0.25 & 0.72/0.33 & 0.91/0.27 \\
DynaSLAM \cite{Bescos2018DynaSLAM} & 0.74/0.26 & 1.57/0.22 & 0.80/0.24 & 0.69/0.18 & 0.45/\textbf{0.09} & \textbf{0.40}/\textbf{0.16} & 0.50/0.17 & 0.52/0.29 & 1.05/0.32 & 0.93/0.29 & 0.67/0.32 & 0.76/0.23 \\
Ours St only & 0.71/\textbf{0.24} & 1.45/0.30 & \textbf{0.75}/0.23 & 0.73/0.19 & 0.47/0.11 & 0.57/0.23 & 0.57/0.22 & 0.51/0.29 & 1.02/0.32 & 0.87/0.26 & 0.65/0.31 & 0.75/0.25 \\
Ours St only (5Hz) & 0.71/0.26 & 1.43/0.23 & 0.78/0.24 & \textbf{0.67}/0.18 & 0.46/\textbf{0.09} & \textbf{0.40}/\textbf{0.16} & \textbf{0.47}/\textbf{0.14} & 0.52/0.29 & 0.99/0.31 & 0.90/0.28 & 0.63/0.31 & \textbf{0.72}/0.22 \\
\hline
\end{tabular}}
\caption{Comparison of camera tracking accuracy - average $t_{rel}$ [\%] and $r_{rel}$ [\si{\degree}/100m] against state-of-the-art stereo and LiDAR SLAM systems. Sequences marked with * contain loops. Note that Stereo-DSO is a purely visual odometry system, so their result is without loop closing. We keep it in the table for completeness. \vspace{-0.5cm}}
\label{tab:traj-lp}
\end{table*}

We perform a quantitative evaluation of our novel prior-based object reconstruction optimisation, using LiDAR input on the KITTI3D Dataset \cite{kitti3d}, comparing with auto-labelling~\cite{Zakharov2020Autolabeling3D}, the most related approach. In addition, we evaluate the camera trajectory errors of our full DSP-SLAM system on both stereo+LiDAR and stereo-only input on the  KITTI Odometry~\cite{kitti_odom} benchmark, comparing with state-of-the-art approaches.  We also provide qualitative results of our full SLAM system on pure monocular input on Freiburg Cars \cite{freiburg-cars} and Redwood-OS \cite{choi2016redwood} Chairs dataset. 

\subsection{\label{det3d}3D Object Reconstruction}

We conduct a quantitative comparison of our object pose estimation on the KITTI3D benchmark,  against auto-labeling~\cite{Zakharov2020Autolabeling3D}, a recent approach to prior-based object shape and pose reconstruction based on image and LiDAR inputs, and using the same shape prior embedding (DeepSDF~\cite{deepsdf}) and similar level of supervision (object masks and sparse depth from the LiDAR measurements).


\noindent\textbf{Experimental Setting:} For a fair comparison, we evaluate our approach using a single image and LiDAR input and take the 2D segmentation masks and initial pose estimates from the  auto-labelling code release~\cite{Zakharov2020Autolabeling3D} as initialization for our own optimization approach. We evaluate the results of pose estimation on the trainval split of KITTI3D which consists of 7481 frames, using  the same metrics  proposed in~\cite{Zakharov2020Autolabeling3D}: BEV AP @ 0.50, 3D AP @ 0.50, and the distance threshold metric (NS) from the nuscenes dataset~\cite{caesar2020nuscenes}.

\noindent\textbf{Results:} We report quantatitive results in Tab.~\ref{tab:object_detection}. Our method achieves better performance under almost all metrics, especially on harder samples. We also visualize the comparison of reconstructed shapes and pose in Fig.~\ref{fig:comparison-auto-labelling}. Auto-labelling~\cite{Zakharov2020Autolabeling3D} does not capture shape accurately for several vehicles: The first two cars on the left side are sedans, but auto-labelling~\cite{Zakharov2020Autolabeling3D} reconstructs them as "beetle"-shaped. In addition, some of the cars on the right side are reconstructed with incorrect poses which do not align with the image. In contrast, DSP-SLAM obtains accurate shape and pose.



\noindent\textbf{Timing Analysis:} To achieve close to real-time performance, we employ a Gauss-Newton solver with faster convergence than first-order methods during our optimization, leading to significant speed-ups.
Tab.~\ref{tab:timing} shows a run-time comparison between a first-order optimizer and our Gauss-Newton solver with analytical gradients. Our method is approximately one order of magnitude faster to complete a single iteration, and requires fewer iterations to converge.
\begin{table}[tbp]
\centering
\begin{tabular}{c|l|c|c}
Method & Energy Terms & msec. / iter & \# of iter \\ \hline
1st order & $E_{surf} + E_{rend}$ & 183 & 50 \\
1st order & $E_{surf}$ & 88 & 50 \\ \hline
Ours GN & $E_{surf} + E_{rend}$ & 20 & 10 \\
Ours GN & $E_{surf}$ & 4 & 10 \\ \hline
\end{tabular}
\caption{Speed comparison between first-order optimization and our Gauss-Newton method with analytical Jacobians\vspace{-0.5cm}}
\label{tab:timing}
\end{table}

\begin{figure*}[htbp]
\centering
\includegraphics[width=1.00\textwidth]{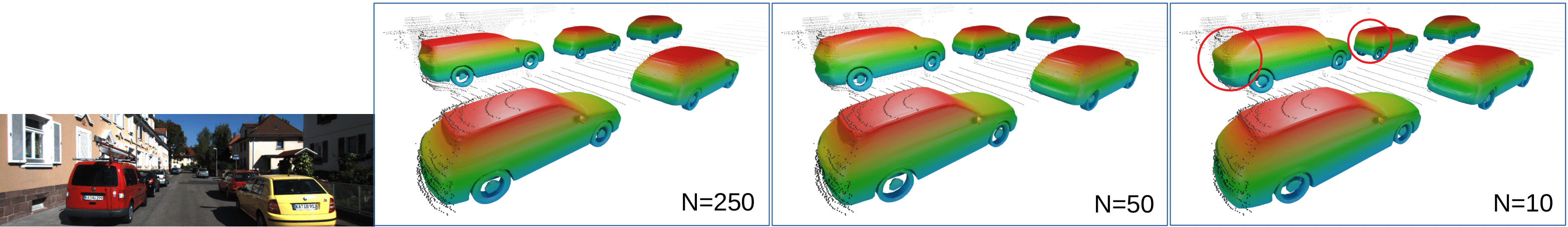}
\caption{Object reconstruction results when using different number of LiDAR points per object (N=250, 50, 10). There is no noticeable difference when the number of points is reduced from 250 to 50. The reconstruction quality starts to degrade when further reducing to 10. The degraded parts are marked with a red circle.
\vspace{-0.1cm}}
\label{fig:recon-num-pts}
\end{figure*}

\noindent\textbf{Ablation Study:} We conducted an ablation study for DSP-SLAM with stereo+LiDAR input to analyse the effect of the number of LiDAR points used for shape optimization  on the reconstruction error.  Fig.~\ref{fig:recon-num-pts} shows that there is no significant difference when reducing the number of LiDAR points from 250 to 50. The reconstruction quality starts to degrade when the number of points is further reduced to 10. 

\subsection{\label{full_slam}KITTI Odometry Benchmark}

We evaluate the camera trajectory error for our full DSP-SLAM system on the KITTI odometry benchmark with both stereo+LiDAR and stereo-only input. We evaluate on the 11 training sequences and compare with state-of-the-art SLAM systems of different input modalities using relative translation error $t_{rel}$ (\%) and relative rotation error $r_{rel}$ (degree per 100m). Quantitative results are shown in Table~\ref{tab:traj-lp}.

\noindent\textbf{Stereo+LiDAR input:} 
The upper part of Tab.~\ref{tab:traj-lp} shows trajectory errors of our system with stereo+LiDAR input. Results suggest our method achieves comparable results with SuMa++, a state-of-the-art LiDAR-based semantic SLAM system~\cite{Chen2019SuMa++}. Note however, that our method only takes very few LiDAR points (several hundred per frame) while SuMa++ uses a full LiDAR point-cloud. It is interesting to see the comparison between our stereo+LiDAR system and stereo ORB-SLAM2, which is used as our backbone system. With our LiDAR-enhanced object reconstruction and joint BA, tracking accuracy improves on most sequences, especially 03, 05, 06, 08 where adequate number of static objects are observed throughout the sequence.
However, our system performs slightly worse on some sequences which contain only moving objects (01, 04) or long trajectory segments where no static objects are observed (02, 10).
The table also shows the effect on the  camera trajectory error when using 250 vs 50 points for object reconstruction. The results suggest that the impact of reducing the number of points on camera tracking accuracy is minimal.

\noindent\textbf{Stereo-only input:} The lower part of Tab.~\ref{tab:traj-lp} contains the results of our stereo-only system. It can be seen that our stereo-only system performs slightly worse than stereo ORB-SLAM2, which means dense shape reconstruction and joint BA does not help improve tracking accuracy with stereo-only input. We argue that the reason is two-fold. Firstly, 3D measurements based on stereo images are noisier than LiDAR-based measurements, giving rise to lower accuracy in object pose estimates.
Secondly, in the stereo-only case, the surface points are obtained from the SLAM system, where the same features are repeatedly measured and not from multiple (LiDAR) measurements.
We also noticed that, to guarantee timings, we were performing bundle-adjustment less frequently than ORB-SLAM2. We re-ran DSP-SLAM, at a slightly reduced frame-rate (5Hz), performing BA after every key-frame (as ORB-SLAM2) and the average performance increased, matching ORB-SLAM2 at $0.72/0.22$. 
A comparison with state-of-the-art stereo SLAM systems is also included in Tab.~\ref{tab:traj-lp}. 


\subsection{Freiburg Cars \& Redwood-OS Dataset}

\begin{figure*}[tbph!]
\centering
\includegraphics[width=0.246\textwidth]{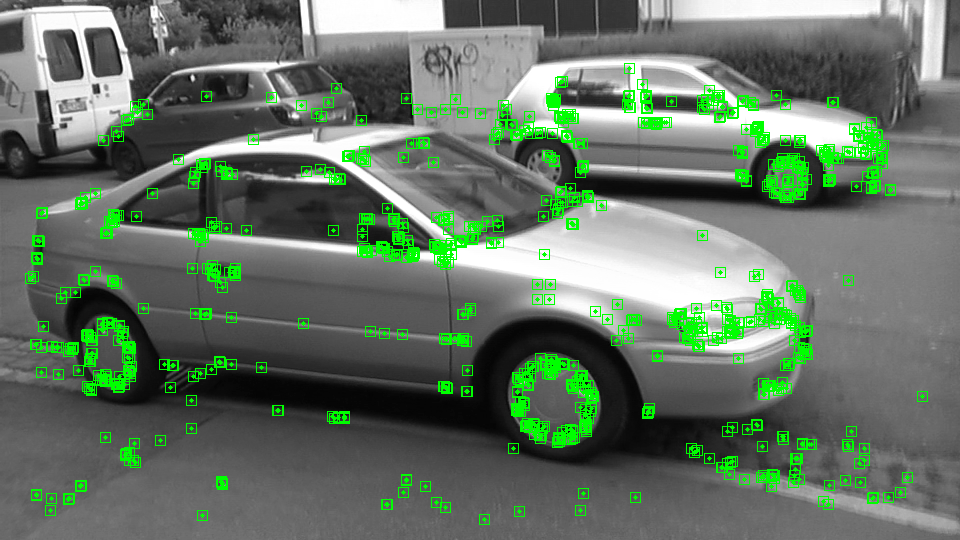}
\includegraphics[width=0.246\textwidth]{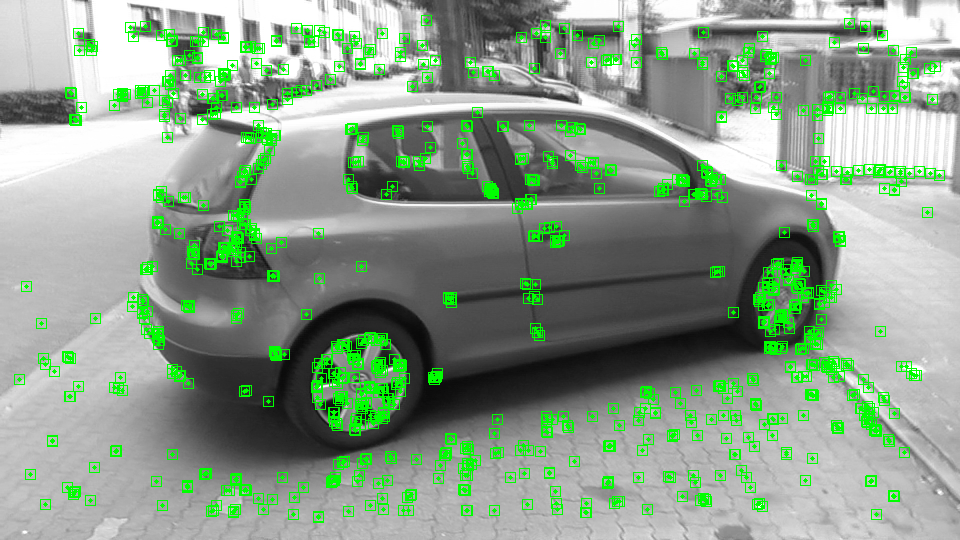}
\includegraphics[width=0.246\textwidth]{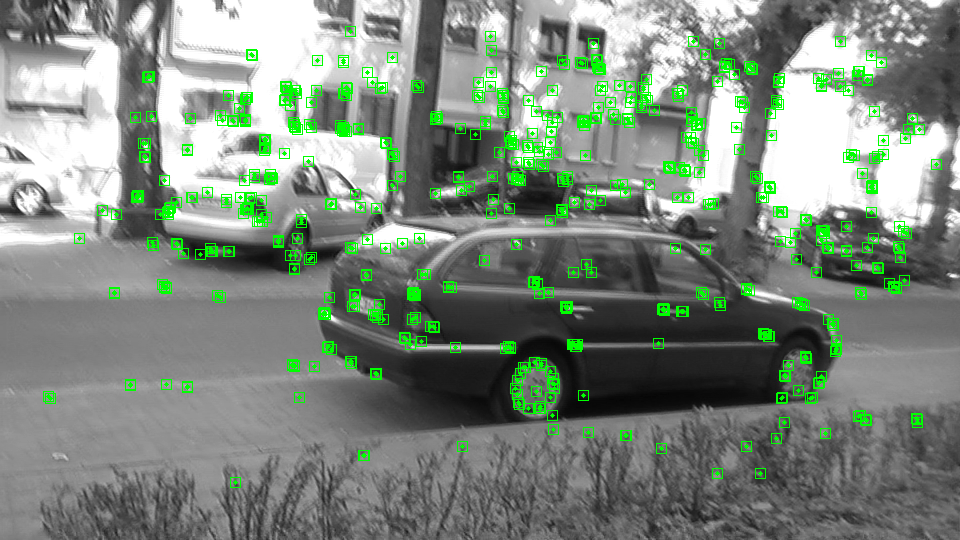}
\includegraphics[width=0.246\textwidth]{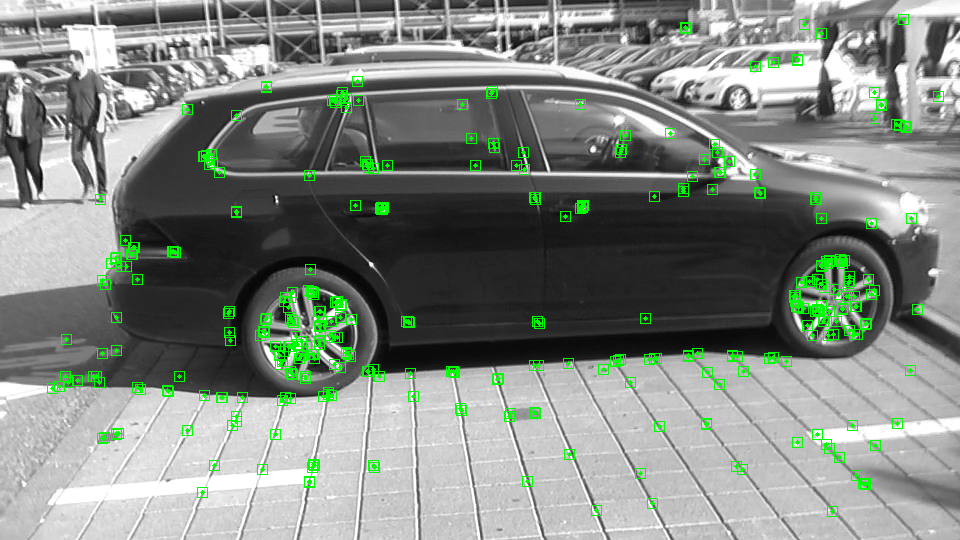}
\includegraphics[width=0.246\textwidth]{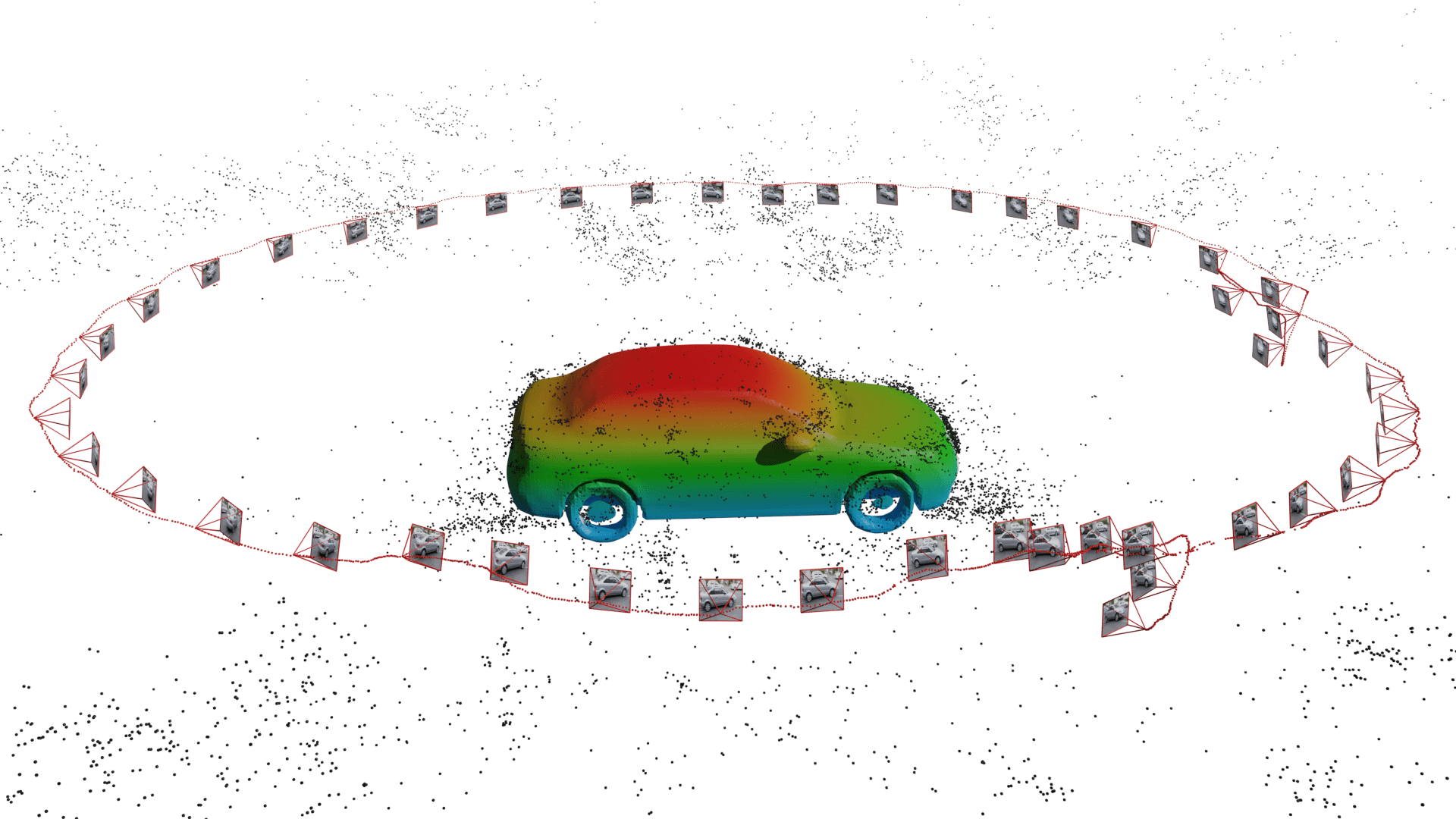}
\includegraphics[width=0.246\textwidth]{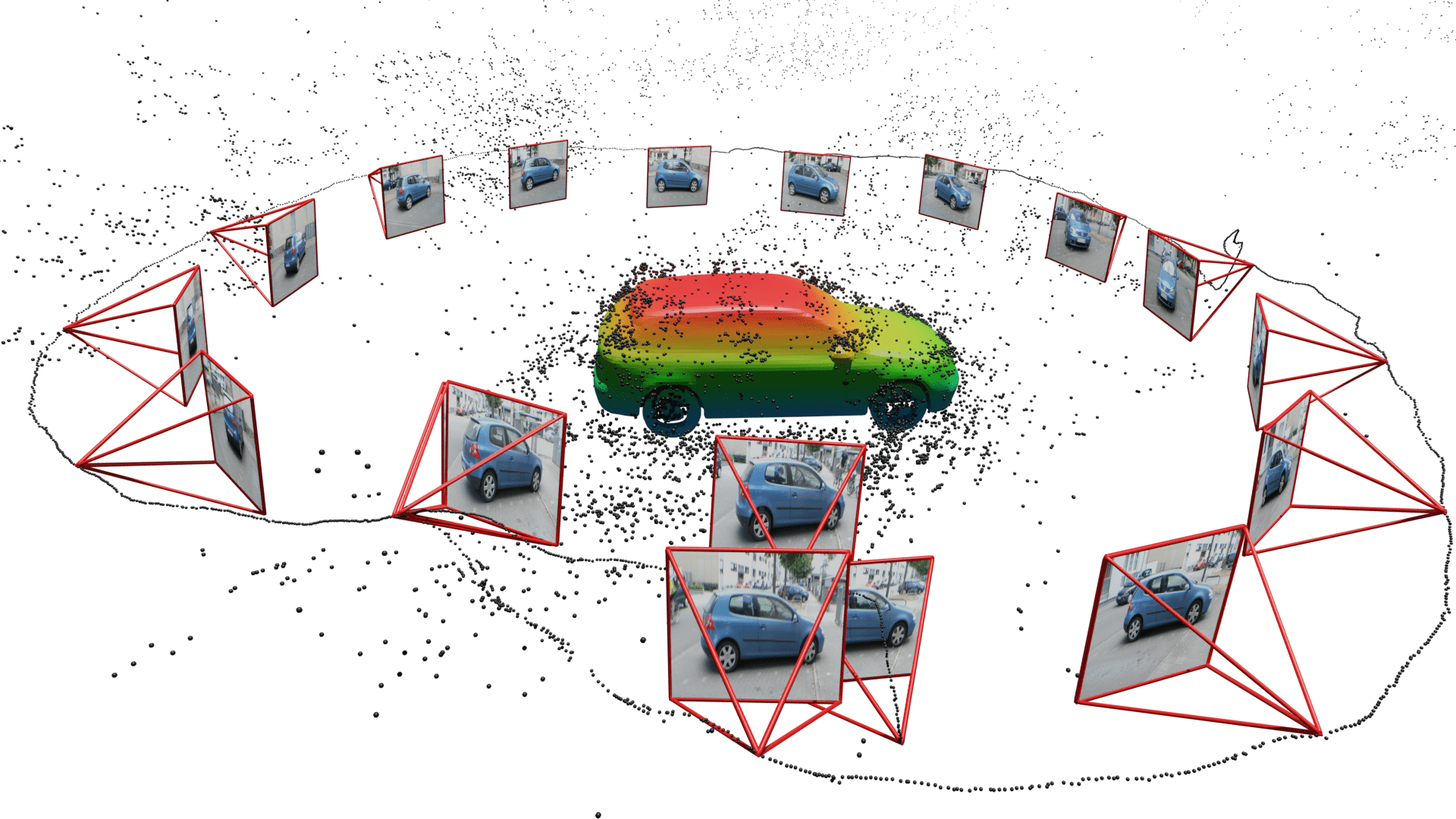}
\includegraphics[width=0.246\textwidth]{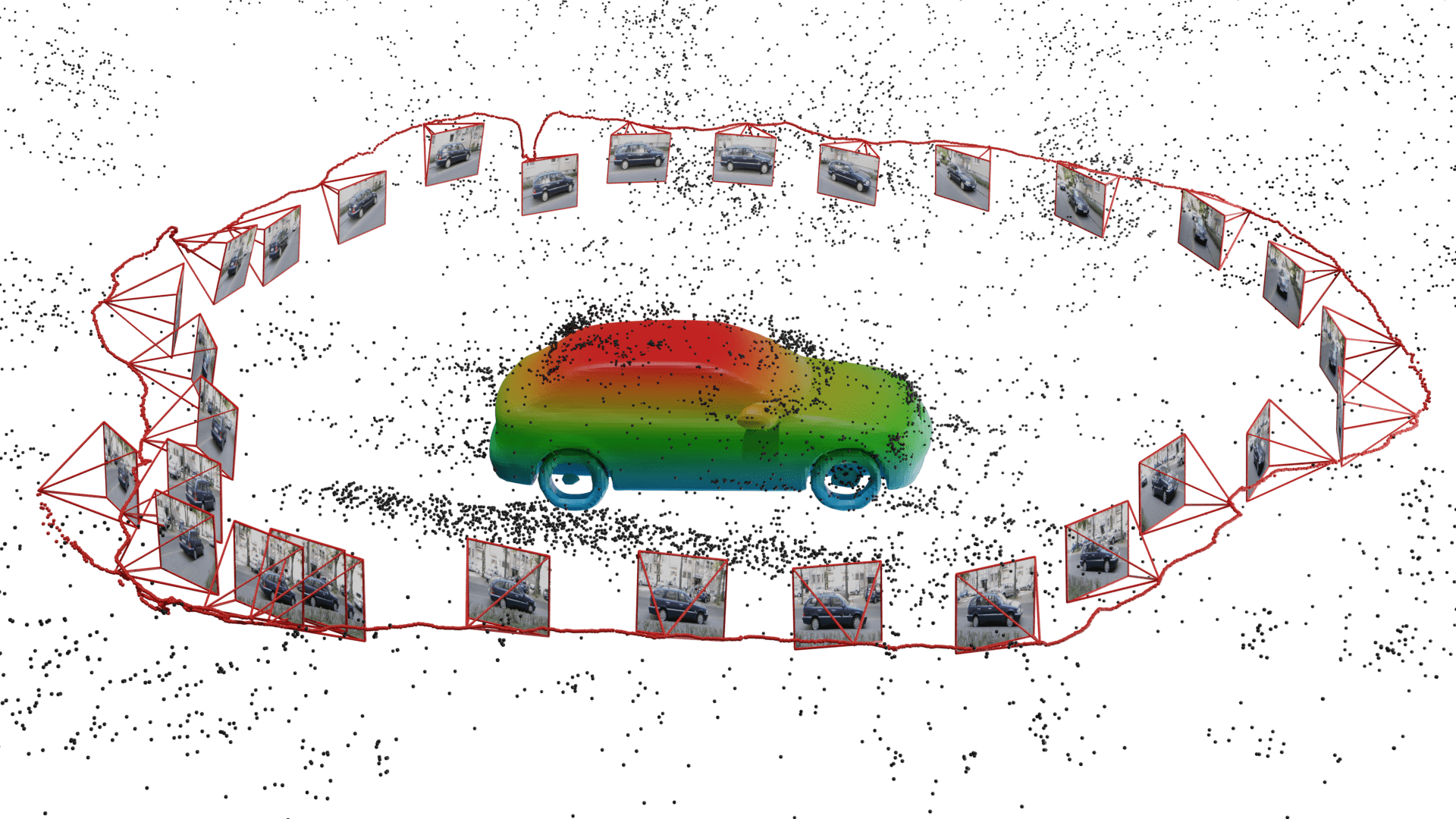}
\includegraphics[width=0.246\textwidth]{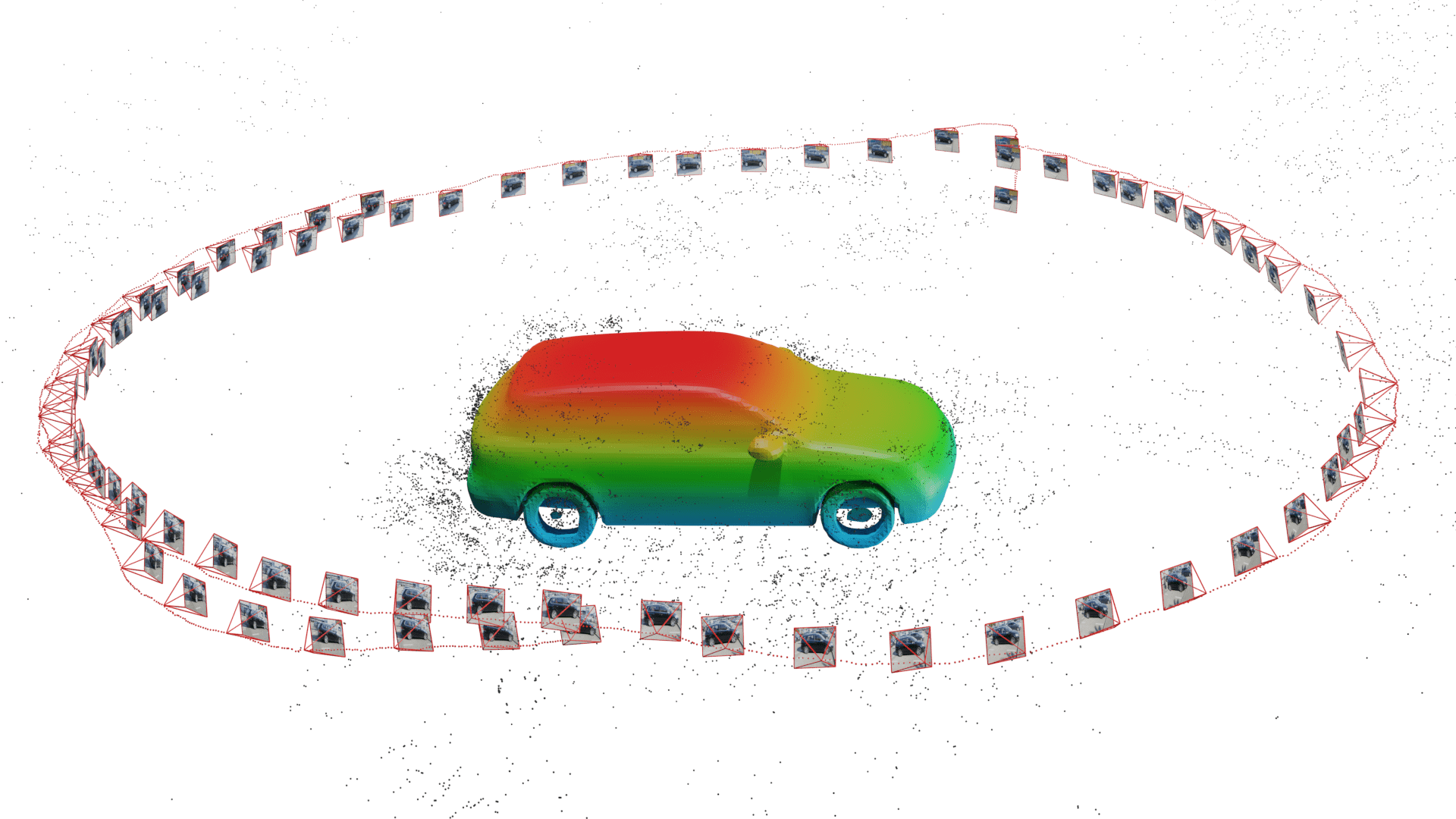}
\caption{Qualitative results on Freiburg Cars dataset
\vspace{-0.1cm}}
\label{fig:freiburg}
\end{figure*}

\begin{figure*}[htbp!]
\centering
\includegraphics[width=0.246\textwidth]{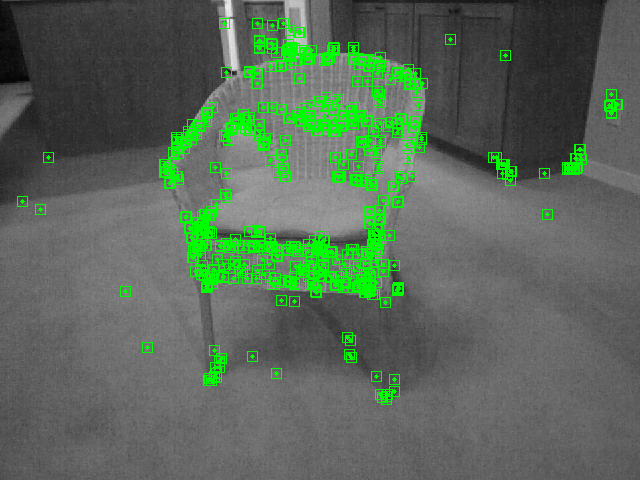}
\includegraphics[width=0.246\textwidth]{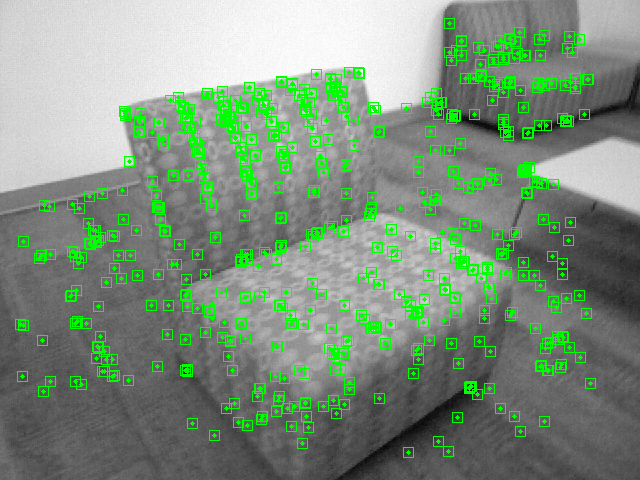}
\includegraphics[width=0.246\textwidth]{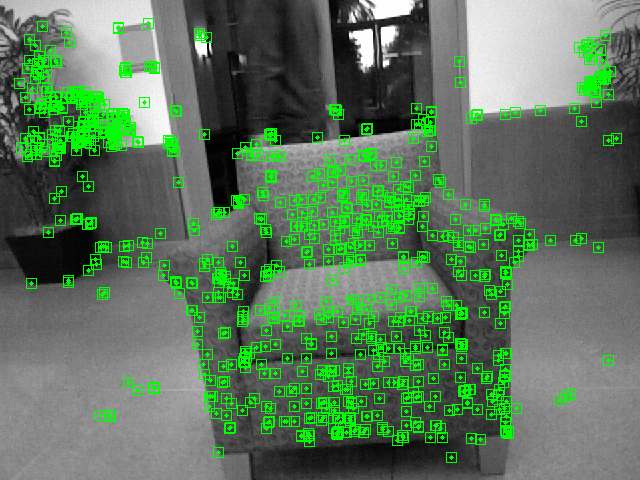}
\includegraphics[width=0.246\textwidth]{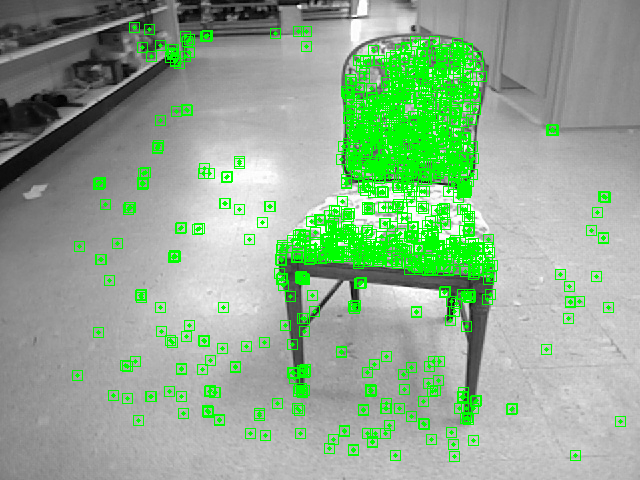}
\includegraphics[width=0.246\textwidth]{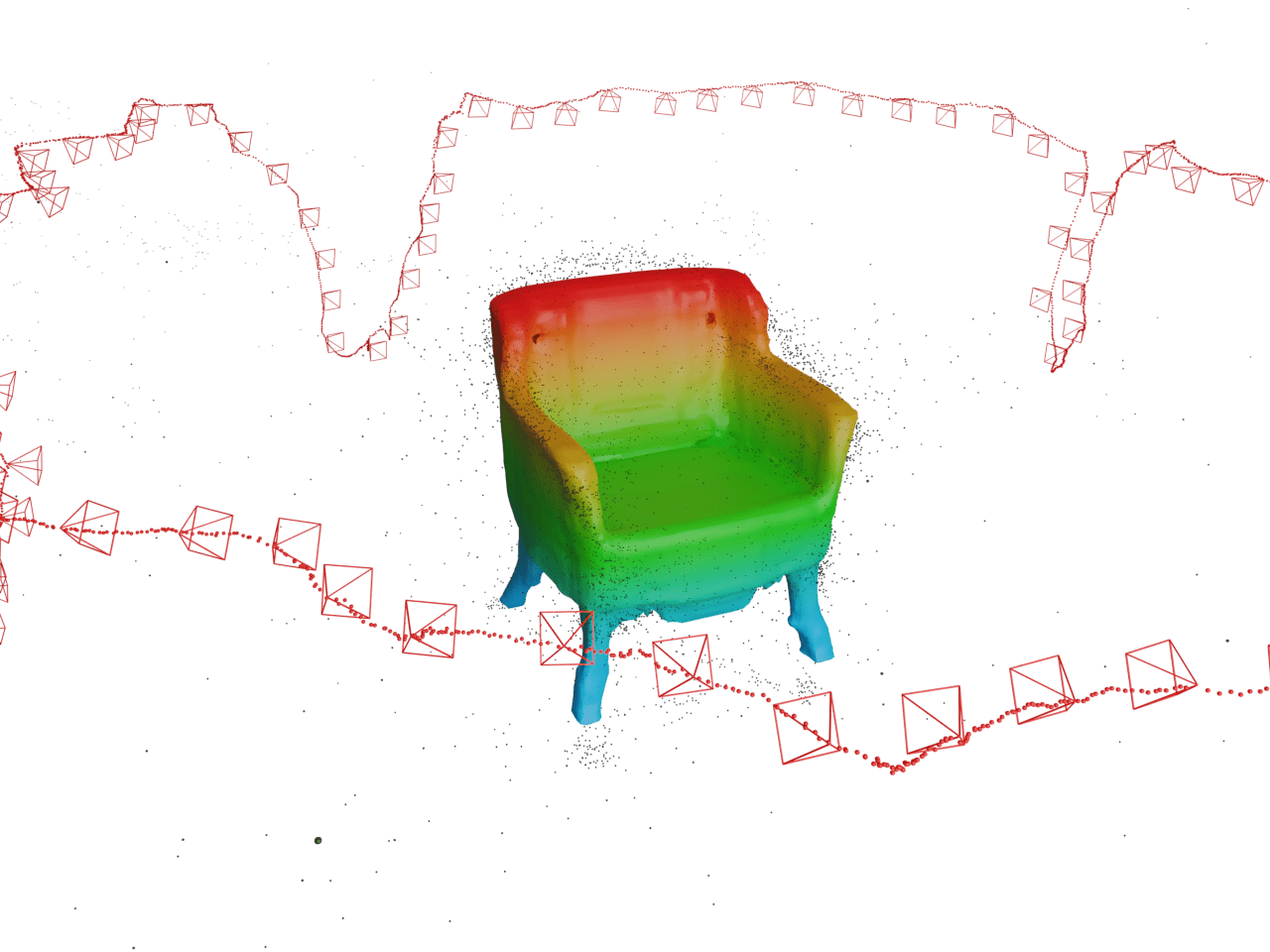}
\includegraphics[width=0.246\textwidth]{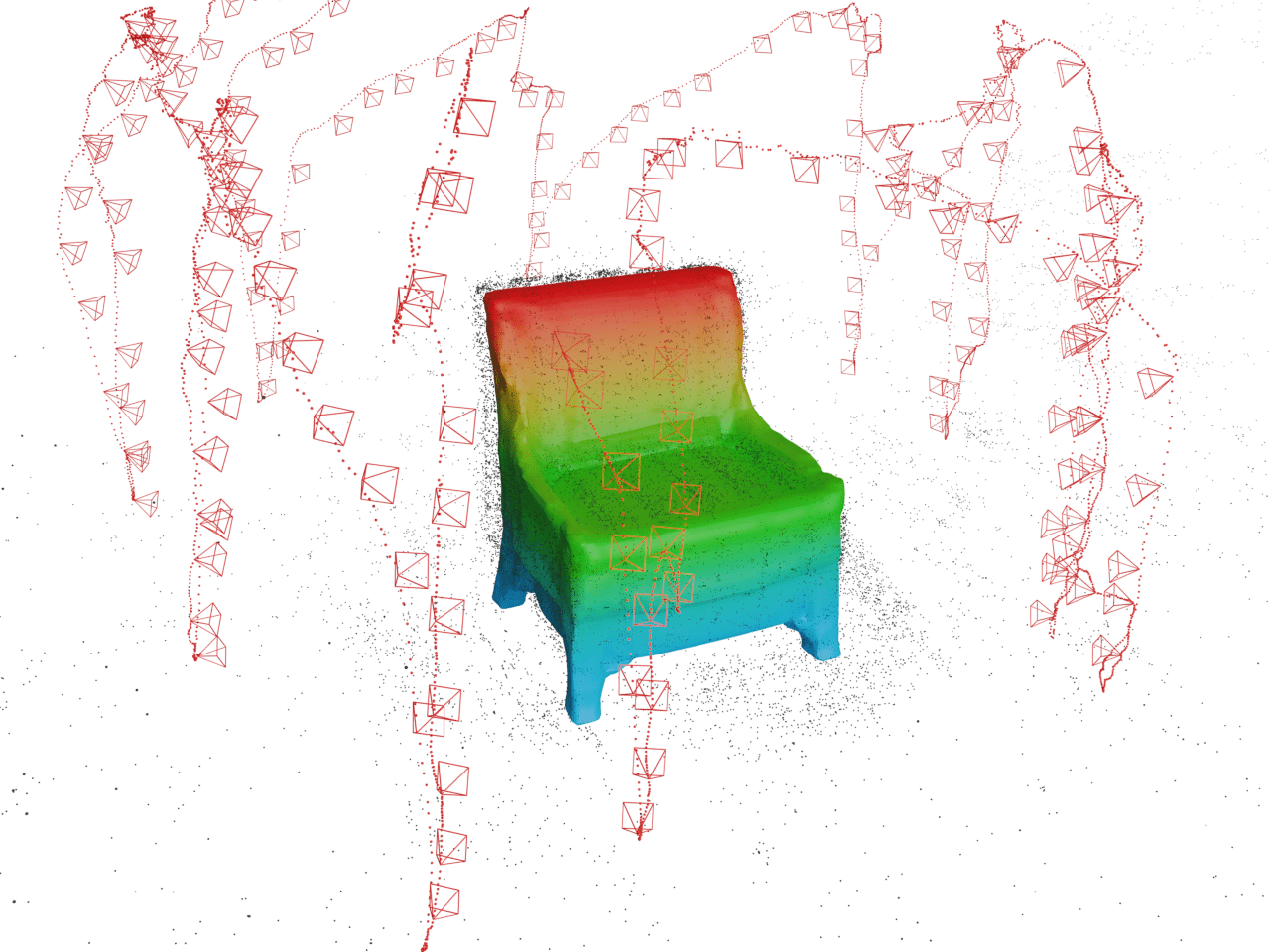}
\includegraphics[width=0.246\textwidth]{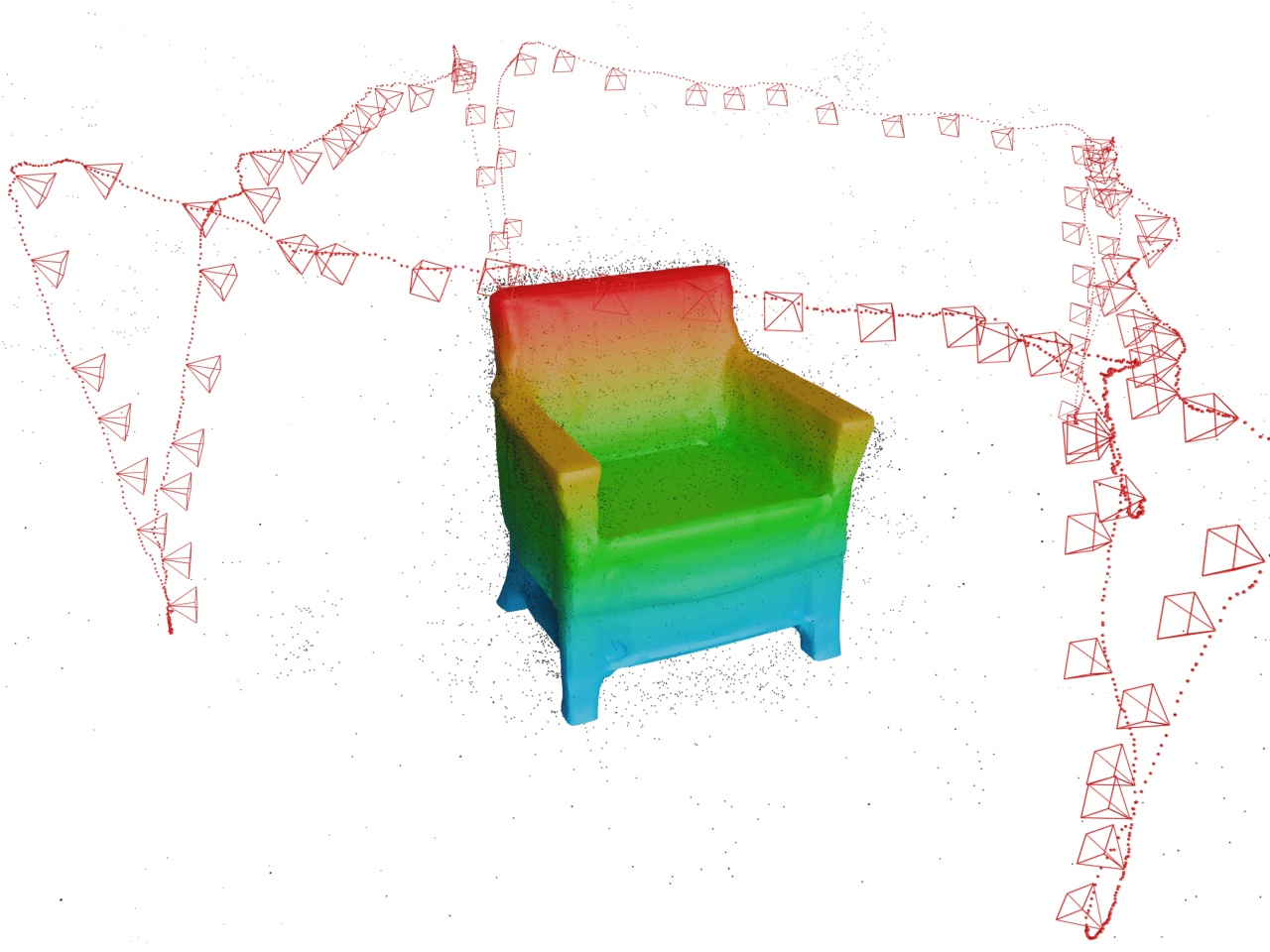}
\includegraphics[width=0.246\textwidth]{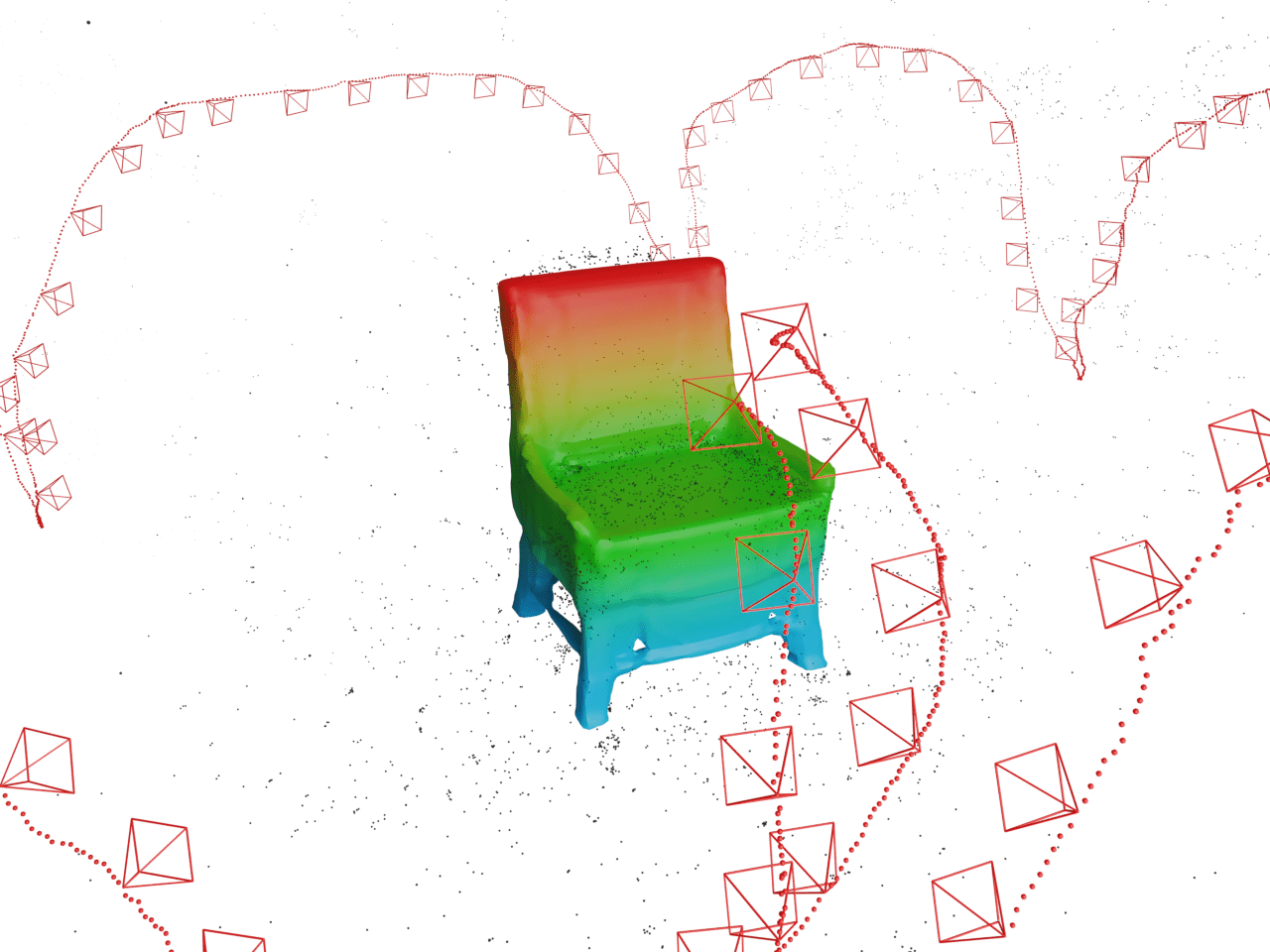}
\caption{Qualitative results on Redwood-OS Chairs dataset
\vspace{-0.4cm}
}
\label{fig:redwood}
\end{figure*}

Finally, we evaluate our SLAM system with monocular input on the Freiburg Cars dataset \cite{freiburg-cars} and Redwood-OS Chairs dataset. Both datasets consist of object-centric sequences with the camera moving around the object. Demonstrations can be seen on Fig.~\ref{fig:freiburg} and~\ref{fig:redwood} and in the supplementary video.

\noindent\textbf{Experimental Setting:} 
3D Bounding boxes are estimated using PCA on the reconstructed surface points. Note that this approach cannot differentiate between the front and back side of the car. To address this issue, we initialize with two flipped hypothesis and keep the orientation that yields a smaller loss.

\noindent\textbf{Results:} Fig.~\ref{fig:freiburg} provides qualitative reconstruction results on 4 Freiburg Cars sequences. Our system is capable of reconstructing dense, accurate and high-quality shapes for cars solely from monocular input at 10-20 fps. Fig.~\ref{fig:redwood} illustrates results on chairs from the Redwood-OS \cite{choi2016redwood} dataset. Reconstruction accuracy is slightly worse than on cars as chairs have more complex shape variations. Results are promising nonetheless -- our method still produces dense meshes that capture the overall object shape from monocular RGB-only sequences, in quasi-real time.

\section{Conclusions}

We have presented DSP-SLAM, a new object-aware real-time SLAM system that exploits deep  shape priors for object reconstruction, produces a joint map of sparse point features for the background and dense shapes for detected objects. We show almost real-time performance on challenging real-world datasets such as KITTI (stereo and stereo+LiDAR), and even on monocular setting Freiburg cars and Redwood-OS. Our quantitative comparisons with competing approaches on camera trajectory estimation and shape/pose reconstruction show comparable or superior performance to state of the art methods.

\vspace{-0.1cm}

\section*{Acknowledgements}

\vspace{-0.1cm}

Research presented here has been supported by the UCL Centre for Doctoral Training in Foundational AI under UKRI grant number EP/S021566/1. We thank Wonbong Jang and Adam Sherwood for fruitful discussions.

\vspace{1cm}

\clearpage

{\small
\bibliographystyle{ieee_fullname}
\bibliography{main}
}

\end{document}


\maketitle


\section{Shape Priors for Object SLAM}

In this section, we explain in more detail the benefit of using learnt shape priors in object SLAM. We evaluate the related object-oriented SLAM methods under three important properties: \textit{1. Can the system reconstruct previously unseen objects? 2. Is the reconstruction complete? 3. Does the reconstruction preserve detailed shape?} 

The first line of works~\cite{SLAM++, Tateno2016When2.5D} relies on a pre-scanned CAD model database to perform online detection and registration. The reconstructed object shapes are complete and detailed, but the system cannot handle new objects. The second category of works~\cite{maskfusion, fusion++} leverages 2D segmentation masks, treat each segmented object as individual identities and perform reconstruction on-the-fly. This reconstruction-by-segmentation strategy can reconstruct arbitrary unseen object shapes in great detail, but the reconstructed shape is incomplete due to partial observation and missing depth. 
The last line of research represents objects as simple geometric shapes, such as ellipsoids~\cite{quadricslam} or cuboids~\cite{cubeslam}. 
This kind of geometric representation can be applied to arbitrary unseen shapes and preserve some important properties like scale and orientation, but we’ve lost the level of details in the shapes.

With learnt shape priors, all the three properties can be achieved at the same time. Its generative property means it can generalize to unseen shapes. Object shape decoded from latent code is guaranteed to be detailed and complete, even from a single partial view. The compact representation also benefits the optimization process. Table \ref{tab:related_work} provides an overview of the related works under the different properties. Only NodeSLAM~\cite{node-slam} and DSP-SLAM can achieve all the three important properties. Unlike NodeSLAM, DSP-SLAM can also deal with large scale environments and monocular RGB inputs.

\section{Full Derivation of Jacobians}

As mentioned in the main paper, one of our contribution is the fast Gauss-Newton optimizer, which is crucial to achieve real-time performance. This section provides full derivation of Jacobians of each individual residual term.

\subsection{Jacobian of Surface Term Residual \label{sec:J_surf}}

As shown in Eq. 1 in the main paper, the surface term residual at pixel $\matr{u} \in \BOmega_s$ is simply the SDF value of the back-projected point under object coordinate:

\begin{equation}
    e_s(\matr{u}) = G(\matr{T}_{oc} \pi^{-1}(\matr{u}, \calD), \matr{z}) = G(^o\matr{x}, \matr{z})
\end{equation}

Its Jacobian with respect to object pose and shape $[\boldsymbol{\xi}_{oc};\matr{z}]$ is defined as:

\begin{equation} \label{eq:J_surf}
    \matr{J}_s = \frac{\partial e_s(\matr{u})}{\partial [\boldsymbol{\xi}_{oc};\matr{z}]}
\end{equation}

where $\boldsymbol{\xi}_{oc} \in \mathbb{R}^7$ is the Cartesian representation (twist coordinate) of the corresponding element in Lie Algebra $\mathfrak{sim}(3)$. The Jacobian with respect to the shape code $\frac{\partial e_s(\matr{u})}{\partial \matr{z}}$ could be obtained directly via back-propagation. To derive the Jacobian with respect to object pose $\boldsymbol{\xi}_{oc}$, we first factorize it using chain rule:

\begin{table}[tbp]
  \center
  \setlength\tabcolsep{4pt}
    \begin{tabular}{c||l|l|l||l|l}
    Method       & \specialcell[b]{Unseen\\Objects}  & \specialcell[b]{Full\\Shape}  & \specialcell[b]{Detailed\\Shape} & \specialcell[b]{RGB\\Input} & \specialcell[b]{Large\\Scene}  \\ \hline \hline
    \specialcell[b]{2.5D is not\\enough\cite{Tateno2016When2.5D}} &  ~ & \checkmark & \checkmark & ~ & ~ \\ \hline
    \specialcell[b]{SLAM++\\\cite{SLAM++}} & ~ & \checkmark & \checkmark & ~ & ~\\ \hline\hline
    \specialcell[b]{Mask-\\Fusion \cite{maskfusion}} & \checkmark & ~ & \checkmark & ~ & ~ \\ \hline
    \specialcell[b]{Fusion++\\\cite{fusion++}} & \checkmark & ~ & \checkmark & ~ & ~\\ \hline\hline
     \specialcell[b]{Quadric-\\SLAM \cite{quadricslam}} & \checkmark & \checkmark & ~ & \checkmark &\checkmark \\ \hline
    \specialcell[b]{Cube-\\SLAM \cite{cubeslam}} & \checkmark & \checkmark & ~ & \checkmark & \checkmark\\ \hline\hline
    \specialcell[b]{Node-\\SLAM \cite{node-slam}} & \checkmark & \checkmark & \checkmark & ~ & ~ \\ \hline
    \textbf{\specialcell[b]{DSP-\\SLAM}} & \checkmark & \checkmark & \checkmark & \checkmark & \checkmark
    \end{tabular}
    \caption{Comparison of the properties of DSP-SLAM with respect
      to other object-oriented SLAM systems. 
      \vspace{-1.0cm}}
    \label{tab:related_work}
\end{table}

\begin{equation} \label{eq:J_surf_pose_1}
    \frac{\partial e_s(\matr{u})}{\partial \boldsymbol{\xi}_{oc}} = \frac{\partial G(^o\matr{x}, \matr{z})}{\partial ^o\matr{x}} \frac{\partial ^o\matr{x}}{\partial \boldsymbol{\xi}_{oc}}
\end{equation}

Then the first term $\frac{\partial G(^o\matr{x}, \matr{z})}{\partial ^o\matr{x}}$ can also be obtained via back-propagation. The second Jacobian term could be computed by applying a left perturbation to the pose $\matr{T}_{oc}$:

\begin{align}
    \frac{\partial ^o\matr{x}}{\partial \boldsymbol{\xi}_{oc}} & = \lim_{\delta \boldsymbol{\xi} = \matr{0}} \frac{\exp{\big(\delta \boldsymbol{\xi}^\wedge}\big) \matr{T}_{oc} {^c\matr{x}} - \matr{T}_{oc} {^c\matr{x}}}{\delta \boldsymbol{\xi}} \\
    & = \lim_{\delta \boldsymbol{\xi} = \matr{0}} \frac{\big(\matr{I} + \delta \boldsymbol{\xi}^\wedge\big) \matr{T}_{oc} {^c\matr{x}} - \matr{T}_{oc} {^c\matr{x}}}{\delta \boldsymbol{\xi}} \\
    & = \lim_{\delta \boldsymbol{\xi} = \matr{0}} \frac{\delta \boldsymbol{\xi}^\wedge \matr{T}_{oc} {^c\matr{x}}}{\delta \boldsymbol{\xi}} = \lim_{\delta \boldsymbol{\xi} = \matr{0}} \frac{\delta \boldsymbol{\xi}^\wedge {^o\matr{x}}}{\delta \boldsymbol{\xi}}  \label{eq:J_surf_pose_2}
\end{align}

where $\exp(\cdot)$ is the exponential map from Lie Algebra $\mathfrak{sim}(3)$ to the corresponding Lie Group $\mathbf{Sim}(3)$, and $\cdot ^\wedge$ is the hat-operator that maps a twist coordinate in $\mathbb{R}^7$ to $\mathfrak{sim}(3)$:

\begin{equation} \label{eq:hat}
    \boldsymbol{\xi}^ \wedge = \begin{bmatrix} \boldsymbol{\nu} \\ \boldsymbol{\phi} \\ s \end{bmatrix}^ \wedge = \begin{bmatrix} \boldsymbol{\phi}_{\times} + s\matr{I} & \boldsymbol{\nu} \\ \matr{0} & 0\end{bmatrix}
\end{equation}

$\boldsymbol{\nu} \in \mathbb{R}^3$, $\boldsymbol{\phi} \in \mathbb{R}^3$ and $s \in \mathbb{R}$ represent the translation, rotation and scale part of the twist coordinate respectively. $(\cdot)_\times$ maps from $\mathbb{R}^3$ to $\mathfrak{so}(3)$ (skew-symmetric matrices). With the equation above, the Jacobian term in Eq. 6 can be computed as: \vspace{-0.3cm}

\begin{align}
    \lim_{\delta \boldsymbol{\xi} = \matr{0}} \frac{\delta \boldsymbol{\xi}^\wedge {^o\matr{x}}}{\delta \boldsymbol{\xi}} & = \lim_{\delta \boldsymbol{\xi} = \matr{0}} \frac{\delta \boldsymbol{\phi}_\times {^o\matr{x}} + \delta s{^o\matr{x}} + \delta \boldsymbol{\nu}}{\delta \boldsymbol{\xi}} \\
    & = \lim_{\delta \boldsymbol{\xi} = \matr{0}} \frac{ \delta \boldsymbol{\nu} - {^o\matr{x}}_\times \delta \boldsymbol{\phi}+ \delta s{^o\matr{x}}}{\delta \boldsymbol{\xi}} \\
    & = \begin{bmatrix} \matr{I} & -^o\matr{x}_{\times} & ^o\matr{x} \end{bmatrix}  \label{eq:J_surf_pose_3}
\end{align}

The full Jacobian of the surface consistency term residual with respect to the object pose can be obtained by combining Eq. \ref{eq:J_surf_pose_1}, \ref{eq:J_surf_pose_2}, \ref{eq:J_surf_pose_3}. Note that here we derive the Jacobian with slight abuse of notations and neglected the notation of homogeneous representation. For a more detailed explanation of Lie Theory, please refer to ~\cite{barfoot_2017} or \cite{sola2018_micro_lie}. 

\subsection{Jacobian of Rendering Term Residual}

As stated in the main paper, the rendering term residual of pixel $\matr{u}$ is

\begin{equation}
    e_r(\matr{u}) = d_\matr{u} - \hat{d}_{\matr{u}}
\end{equation}

To compute the Jacobian of the rendering terms $\matr{J}_{r}$, we can expand it using chain rule:

\begin{align} 
    \matr{J}_{r} & = \frac{\partial e_{r}}{\partial \left[\xibf_{oc}; \matr{z} \right]} \\
    & = \sum_{k=1}^M \frac{\partial e_{r}}{\partial o_k} \frac{\partial o_k}{\partial s_k} \frac{\partial{G({^o\matr{x}_k}, \matr{z})}}{\partial \left[ \xibf_{oc}; \matr{z} \right]} \label{eq:J_render}
\end{align}

\begin{figure}[tbp]
\centering
\includegraphics[width=0.48\textwidth]{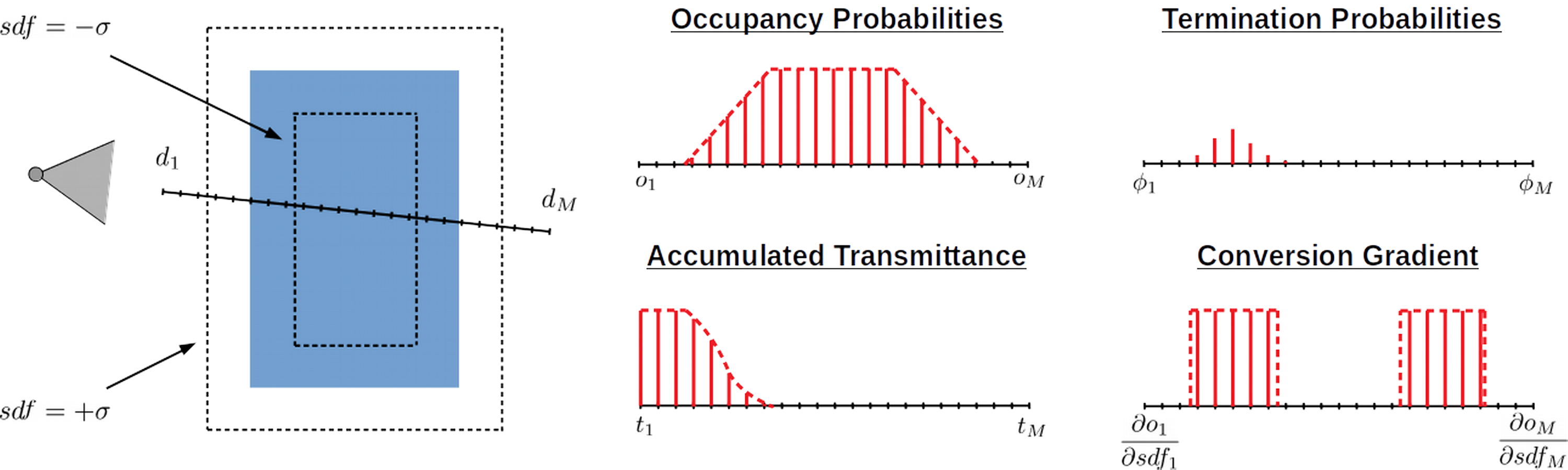}
\caption{Demonstration of the depth rendering process. Only very few ray points contribute to the overall Jacobian term.
\vspace{-0.5cm}}
\label{fig:render_values}
\end{figure}

where $\{^o\matr{x}_k\}_{k=1}^M$ is the depth-ordered set of sampled points along the ray of back-projecting pixel $\matr{u}$. The third term $\frac{\partial{G( ^o\matr{x}_k, \matr{z})}}{\partial \left[ \boldsymbol{\xi}_{oc}; \matr{z} \right]}$ has exactly the same form as the surface term Jacobian in Eq.~\ref{eq:J_surf}, thus it can be computed following Sec.~\ref{sec:J_surf} . The second term $\frac{\partial o_k}{\partial s_k}$ is either a constant value or $0$, as the sdf-occupancy conversion is linear inside the cut-off threshold and constant elsewhere. 

\begin{equation} \label{eq:conversion_grad}
    \frac{\partial o_k}{\partial s_k} =   
    \begin{cases}
        -\frac{1}{2\sigma} & |s_k| < \sigma\\
        0 & \text{elsewhere}
    \end{cases}
\end{equation}

To compute the first term, we expand the residual term using Eq. 3 and Eq. 4 in the main paper following~\cite{multi-view_supervision}: \vspace{-0.3cm}

\begin{equation*}
    \begin{aligned}
        e_r & = d_\matr{u} - \bigg( \sum_{i=1}^M d_i \ o_i \prod_{j=1}^{i-1} (1-o_j) + d_{M+1} \prod_{j=1}^M (1 - o_j) \bigg) \\
        & = \sum_{i=1}^M \psi(i) \ o_i \prod_{j=1}^{i-1} (1-o_j) + \psi(M+1) \prod_{j=1}^M (1 - o_j) \\
        & = \sum_{i=1}^{M+1} \psi(i) \prod_{j=1}^{i-1} (1-o_j) - \sum_{i=1}^M \psi(i) \prod_{j=1}^i (1-o_j) \\
        & = \psi(1) + \sum_{i=1}^M (\psi(i+1) - \psi(i)) \prod_{j=1}^i (1-o_j)
    \end{aligned}
\end{equation*}

where $\psi(i) = d_{\matr{u}} - d_i$ is the depth residual for each of the sampled points along the ray. Differentiating with respect to the sample point, we reach: \vspace{-0.2cm}

\begin{align} 
    \frac{\partial e_{r}}{\partial o_k} 
    & = \sum_{i=1}^M (\psi(i+1) - \psi(i)) \frac{\partial}{\partial o_k}\prod_{j=1}^i (1-o_j) \\
    & = \sum_{i=k}^M (\psi(i) - \psi(i+1)) \prod_{j=1, j \neq k}^i (1-o_j) \\
    & = \Delta d\sum_{i=k}^M \prod_{j=1, j \neq k}^i (1-o_j) \label{eq:dr_do}
\end{align}

As we are sampling uniformly along the ray, the coefficient $\psi(i) - \psi(i+1) = d_{i+1} - d_i$ reduces to $\Delta d$. To understand this expression we can multiply $(1-o_k)$ on both sides, of Eq.~\ref{eq:dr_do}, which gives us:

\begin{equation}
        \frac{\partial e_{r}}{\partial o_k}(1 - o_k)  = \Delta d\sum_{i=k}^M t_i
\end{equation}

where $t_i = \prod_{j=1}^i (1 - o_j)$ represents the accumulated transmittance at point $^o\matr{x}_i$ along the ray. Before hitting the surface, $o_k$ is zero, so the term $(1-o_k)$ has no effect, and the Jacobian term becomes the sum of transmittance of all the points starting from point k. And after the ray entering the solid body, this Jacobian term will be zero. This means only points before entering the solid body contribute to the rendered depth value.

Now from Eq.~\ref{eq:conversion_grad} and Eq.~\ref{eq:dr_do}, we already know that many Jacobian terms that make up the final Jacobian in Eq.~\ref{eq:J_render} should be zero. Therefore, we could further speed up the optimization by not evaluating the third term $\frac{\partial{G( ^o\matr{x}_k, \matr{z})}}{\partial \left[ \boldsymbol{\xi}_{oc}; \matr{z} \right]}$ at those points. Figure~\ref{fig:render_values} is a demonstration of the rendering process and the sparse nature of the resulting Jacobians.

\subsection{Jacobian of Prior Terms}

Based on the shape regularisation energy defined in Eq. 6 in the main paper. The residual of this energy term is simply the shape code vector itself:

\begin{equation} \label{eq:E_code}
    e_c = \matr{z}
\end{equation}

Based on Eq.~\ref{eq:E_code}, we have

\begin{equation}
    \frac{\partial e_c}{\partial \xibf_{oc}} = \matr{0}, \frac{\partial e_c}{\partial \matr{z}} = \matr{I}
\end{equation}

Optionally, we can also apply structural priors such as constraining the object orientation to be aligned with the ground plane, as in~\cite{wang2020directshape}. We can define the rotation prior residual as

\begin{align}
    e_{rot} & = 1 - \matr{R}_{co}(0:2,1) \cdot \matr{n}_g \\
    & = 1 - \begin{bmatrix} 0 & 1 & 0 \end{bmatrix}\matr{R}_{oc}\matr{n}_g
\end{align}

\begin{figure}[tpb]
\centering
\includegraphics[width=0.48\linewidth]{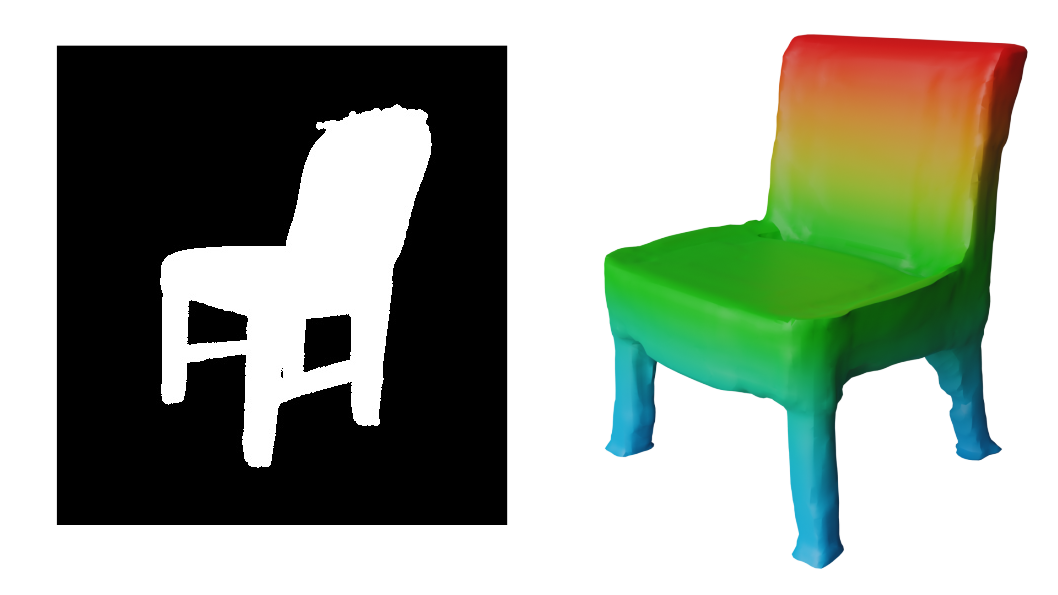}
\includegraphics[width=0.48\linewidth]{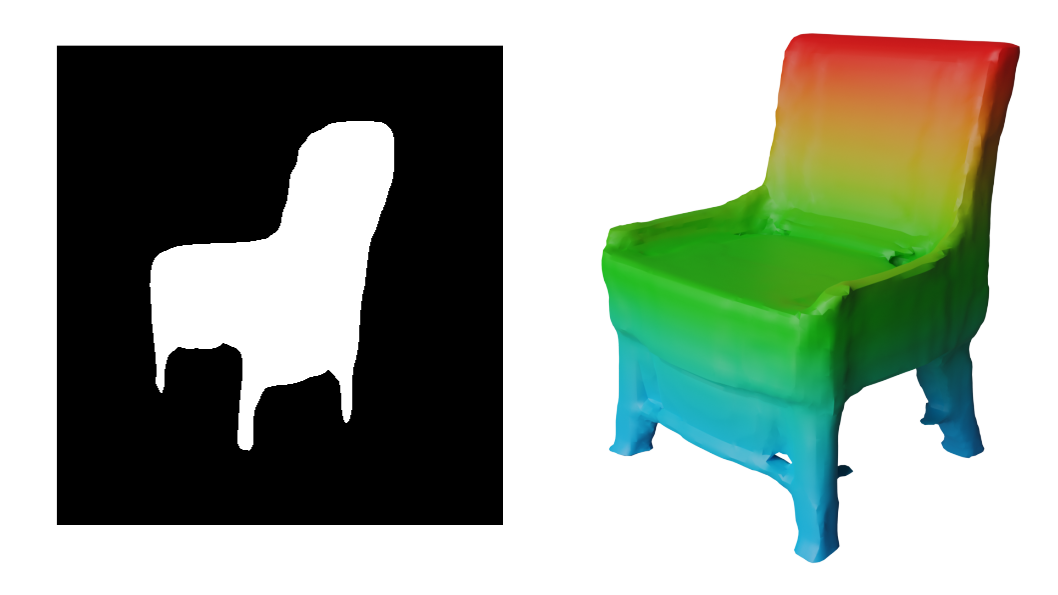}
\caption{Results with GT ($\leftarrow$) vs. MaskRCNN ($\rightarrow$) masks \vspace{-0.6cm}}
\label{fig:gt_mask}
\end{figure}


where $\matr{R} \in \mathbf{SO}(3)$ denotes the rotation part of the transformation matrix, and $(0:2, 1)$ represents the operation of getting the second column of the matrix. $\matr{n}_g$ is the normal direction to the ground plane under camera coordinate frame. The Jacobian with respect to pose can be easily obtained following Eq.~\ref{eq:J_surf_pose_3}:

\begin{align}
    \frac{\partial e_{rot}}{\partial \xibf_{oc}} &= - \begin{bmatrix} 0 & 1 & 0 \end{bmatrix}\frac{\partial \matr{R}_{oc}\matr{n}_g}{\partial \xibf_{oc}} \\
    & = \begin{bmatrix} 0 & 1 & 0 \end{bmatrix} \begin{bmatrix} \matr{0} & (\matr{R}_{oc}\matr{n}_g)_{\times} & \matr{0} \end{bmatrix} \\
    & = \begin{bmatrix} 0 & 0 & 0 & (\matr{R}_{oc}\matr{n}_g)_{\times}(1,0:2) & 0 \end{bmatrix}
\end{align}

where $(0:2, 1)$ represents the operation of getting the second row of the matrix. As this residual has no effect on object shape, we have:

\begin{equation}
    \frac{\partial e_{rot}}{\partial \matr{z}} = \matr{0}
\end{equation}

\section{Experiment Details and Run-time Analysis}

We evaluate the run-time performance of our full SLAM system (stereo+LiDAR) on a Linux system with Intel Core i9 9900K CPU at 3.60GHz, and an nVidia GeForce RTX2080 GPU with 8GB of memory. The 2D detection boxes and masks are obtained using MaskRCNN~\cite{maskrcnn},\footnote{\url{https://github.com/facebookresearch/detectron2}} throughout all the experiments. Initial object poses are inferred using the LiDAR-based 3D b-box detector SECOND~\cite{second}.\footnote{\url{https://github.com/traveller59/second.pytorch}} Our object reconstruction pipeline is implemented in Python with PyTorch. The back-end object-SLAM framework is implemented in C++ as an extension of the original ORB-SLAM2 implementation. The Python part is embedded into C++ using pybind11. 

Table~\ref{tab:timing} lists the breakdown of the run-time performance of different major components. Note that all those components are performed only at key-frames. 
For KITTI sequences, we adopted two design choices to achieve real-time performance. Firstly, we noticed shape estimation did not improve much over time in KITTI, thus we perform full optimization only for new objects and only update poses for existing objects (as stated in the main paper). Secondly, we abort new object optimization whenever necessary to ensure new key-frame is inserted in time. These make our system able to operate at $10$Hz on KITTI. In sequences such as Freiburg and Redwood-OS, it is beneficial to update the shape more frequently, with fewer GN iterations to guarantee real-time performance. Object meshes can be extracted optionally using marching cube with an extra GPU, for visualization only.

\begin{table}[tbp]
\centering
\begin{tabular}{l|c}
\textbf{Component} & \textbf{Time (ms)} \\ \hline
Mask R-CNN Detector & 70 / frame \\
3D LiDAR Detector & 60 / frame \\
Pose-shape Optimization & 20$\times$10 / new object  \\
Pose-only Optimization & 4$\times$5 / vis. object  \\ \hline
\end{tabular}
\caption{Run-time analysis with system components \vspace{-0.4cm}}
\label{tab:timing}
\end{table}


\section{Ablation with GT masks}
We have shown promising reconstruction results on cars on both KITTI and Freiburg dataset. Chairs have thin structures and a complex topology and are more challenging to reconstruct than cars. We noticed the thin legs of the last chair in Fig.~10 in the main paper were not properly reconstructed. This is because our reconstruction makes use of Mask-RCNN segmentation masks, which are noisy and affected by shadows. This can result in background points being associated to the chair, leading to incorrect results. We conducted ablation study using ground-truth masks as input. Fig.~\ref{fig:gt_mask} illustrates an upper bound quality that could be achieved with clean GT masks.

\section{More Qualitative Results}

\begin{figure*}[tpb]
\centering
\includegraphics[width=0.48\textwidth]{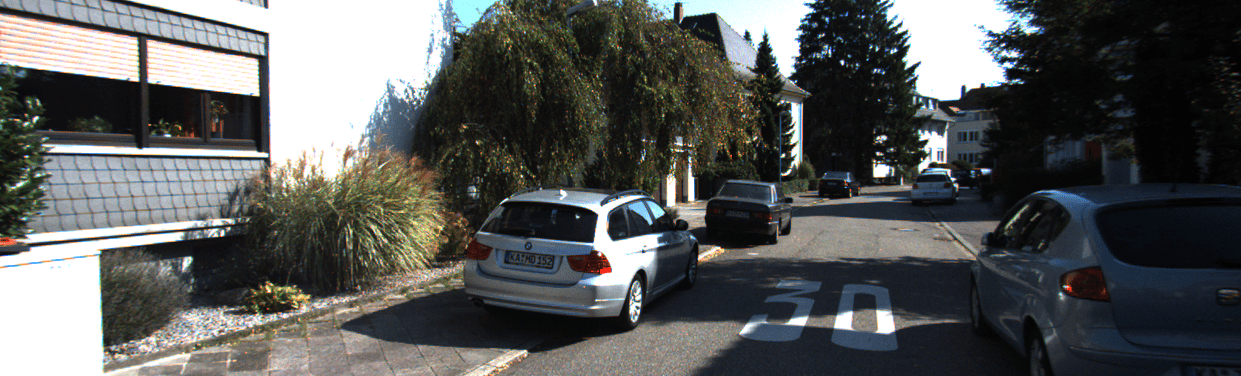}
\includegraphics[width=0.48\textwidth]{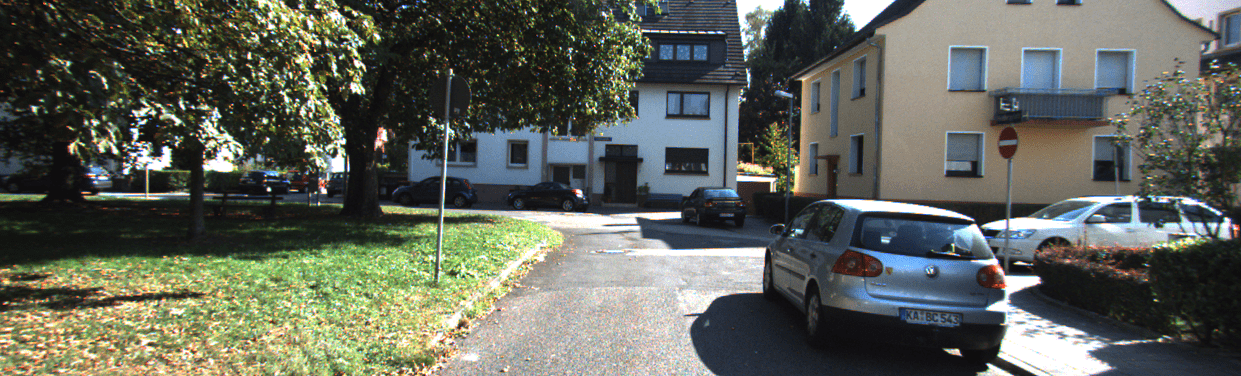}
\includegraphics[width=0.48\textwidth]{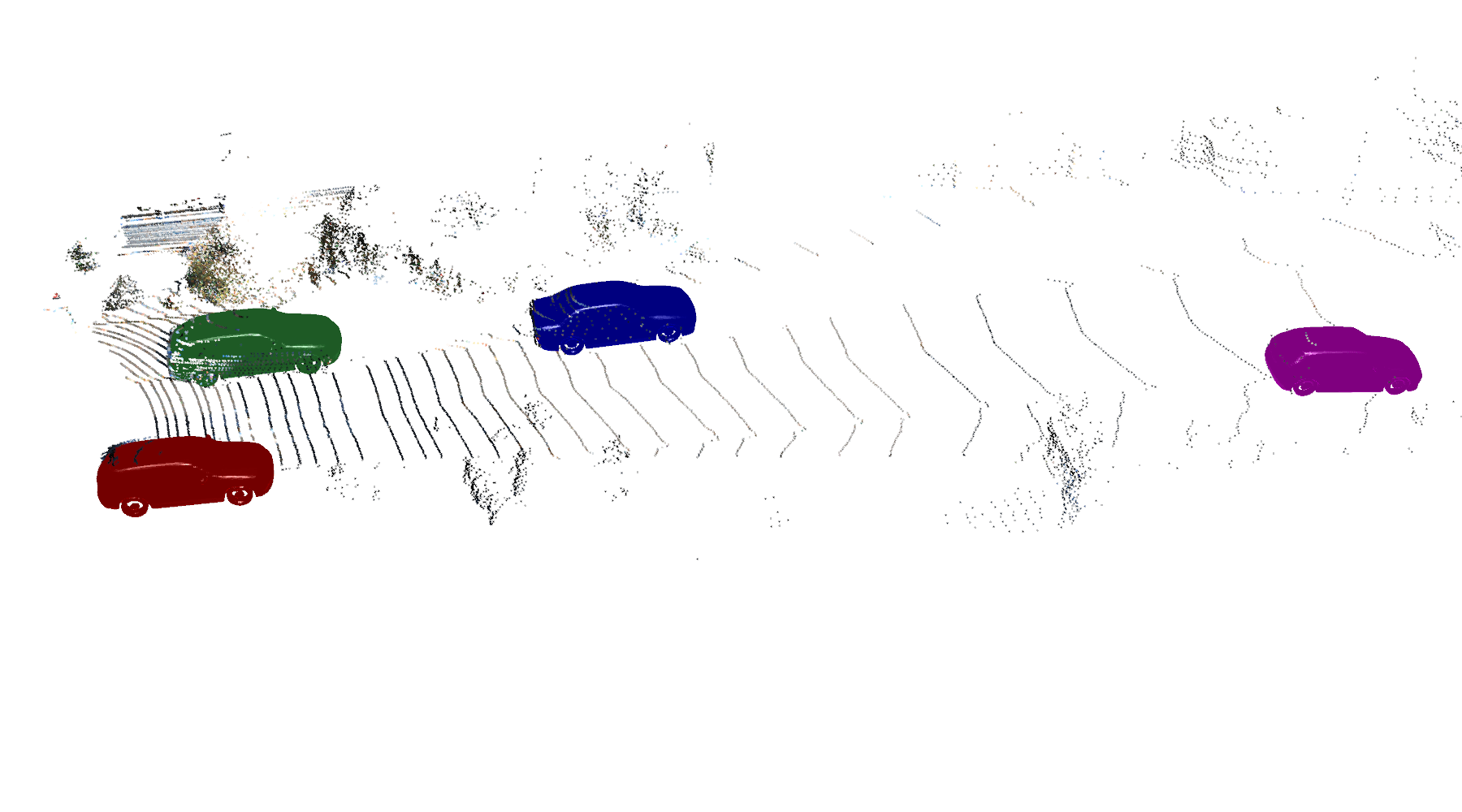}
\includegraphics[width=0.48\textwidth]{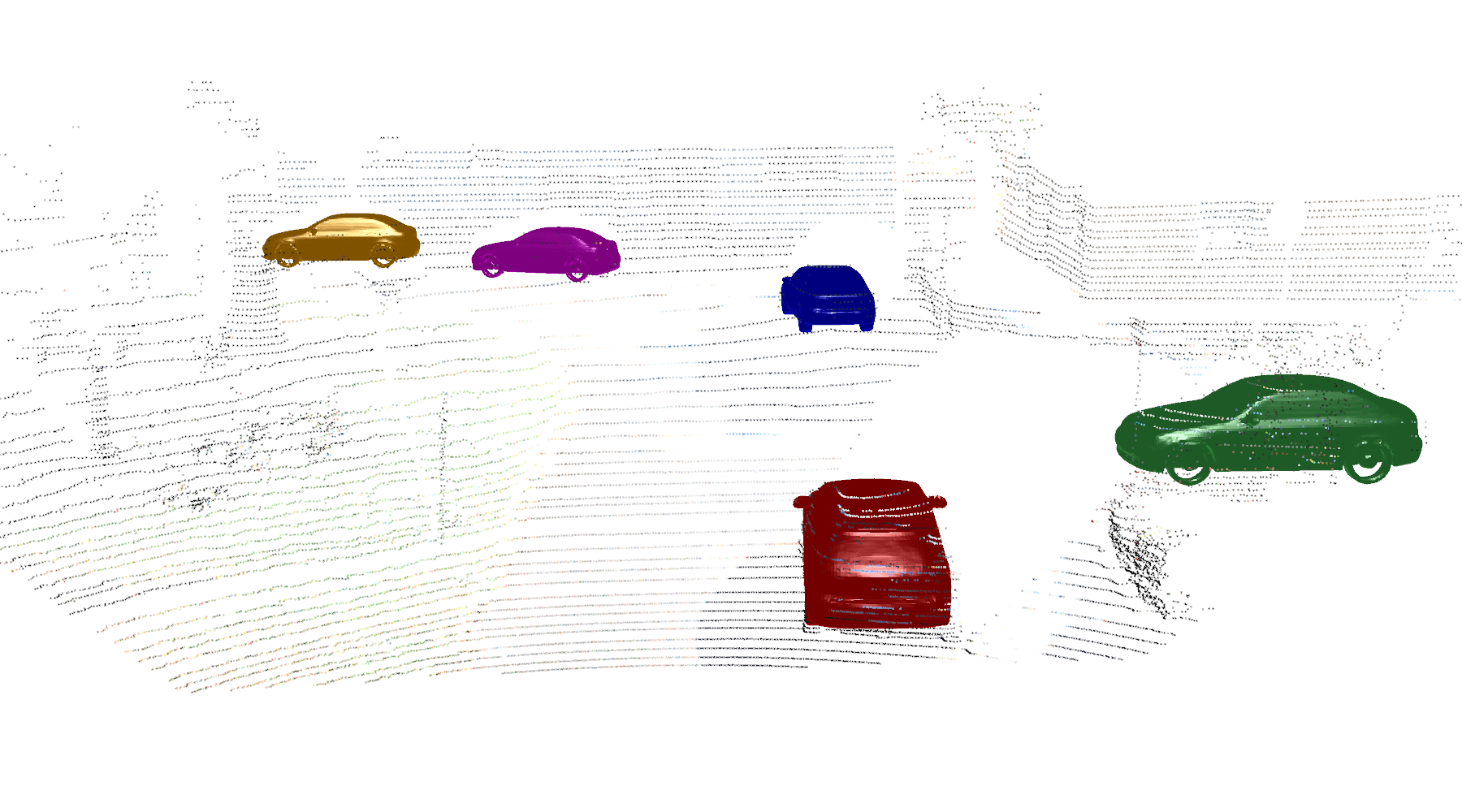}
\includegraphics[width=0.48\textwidth]{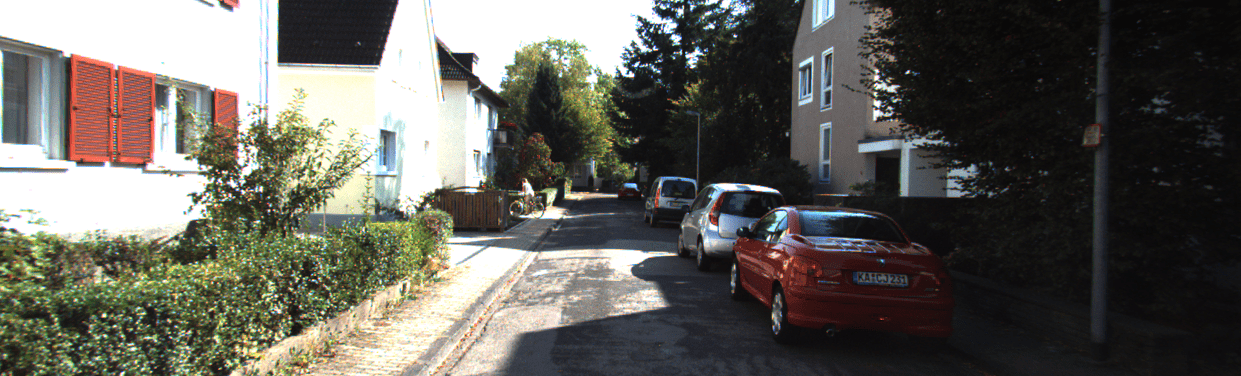}
\includegraphics[width=0.48\textwidth]{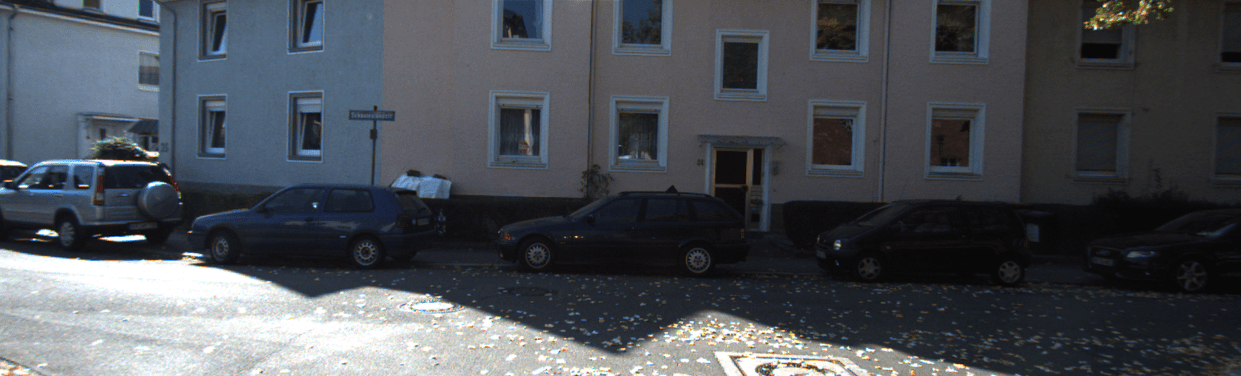}
\includegraphics[width=0.48\textwidth]{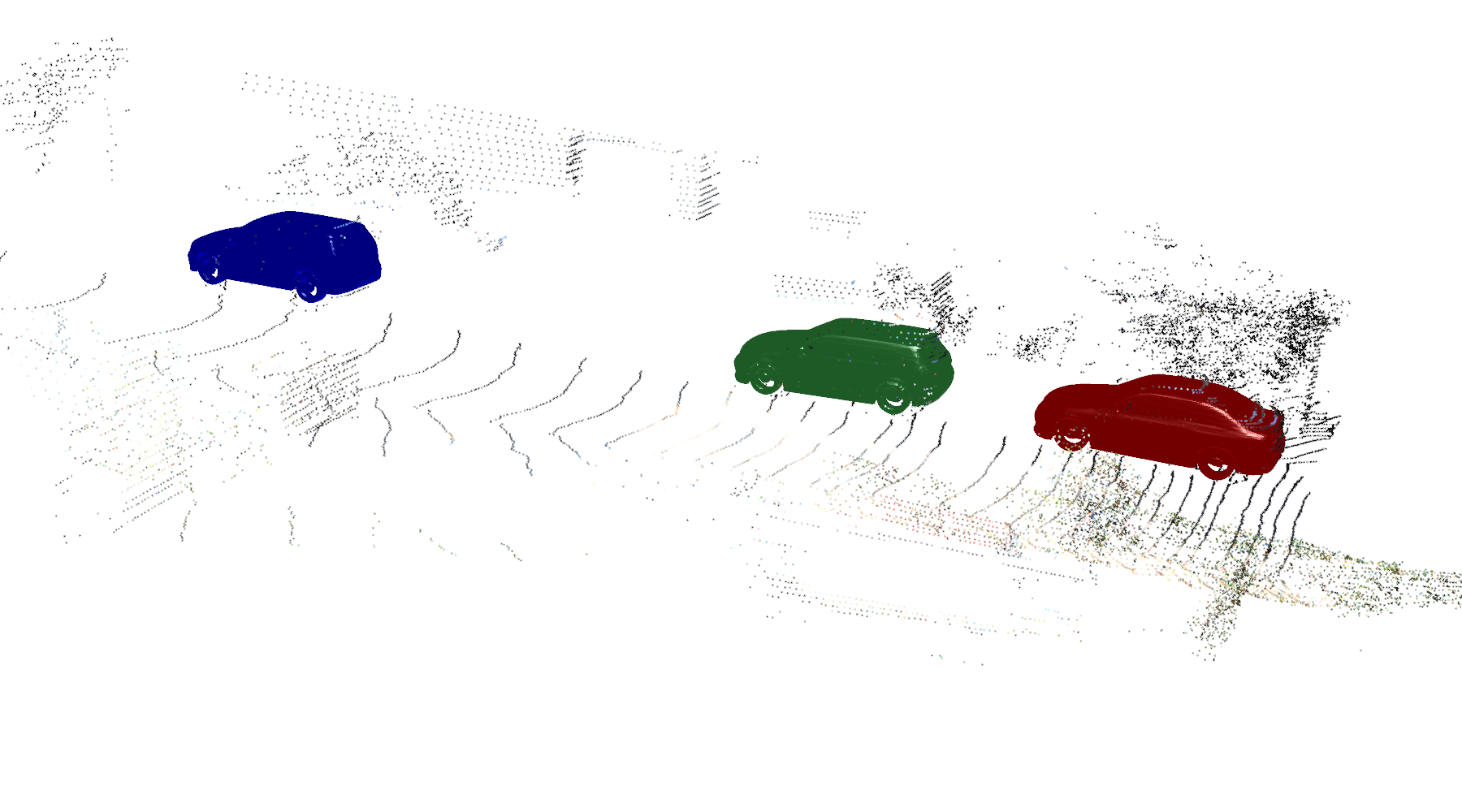}
\includegraphics[width=0.48\textwidth]{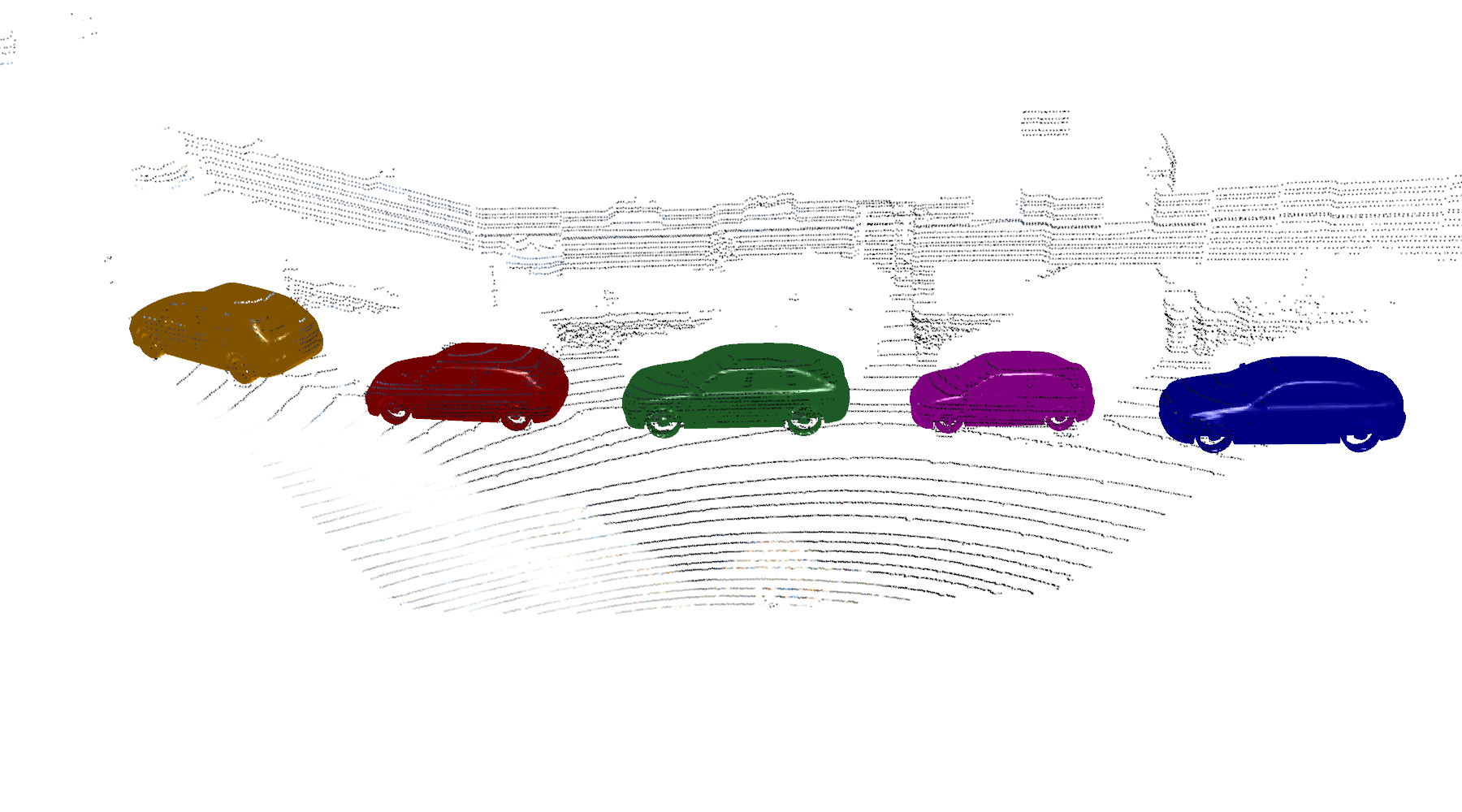}
\includegraphics[width=0.48\textwidth]{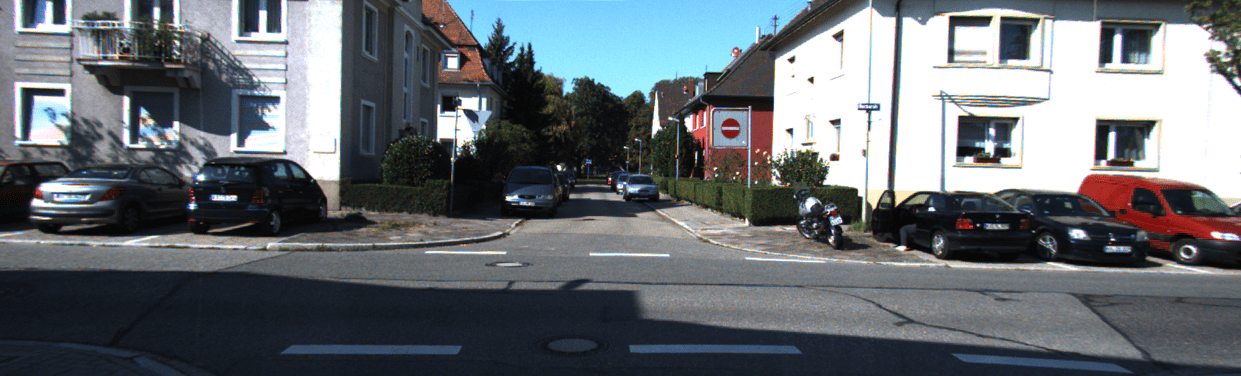}
\includegraphics[width=0.48\textwidth]{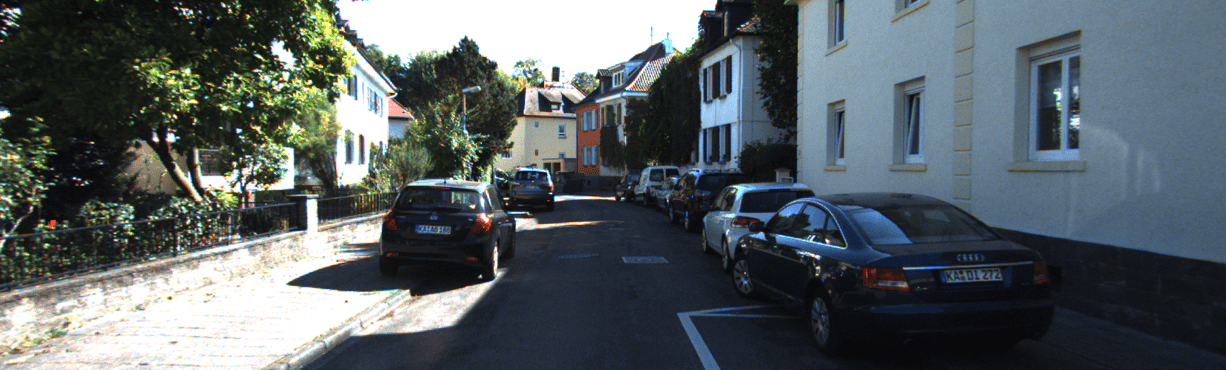}
\includegraphics[width=0.48\textwidth]{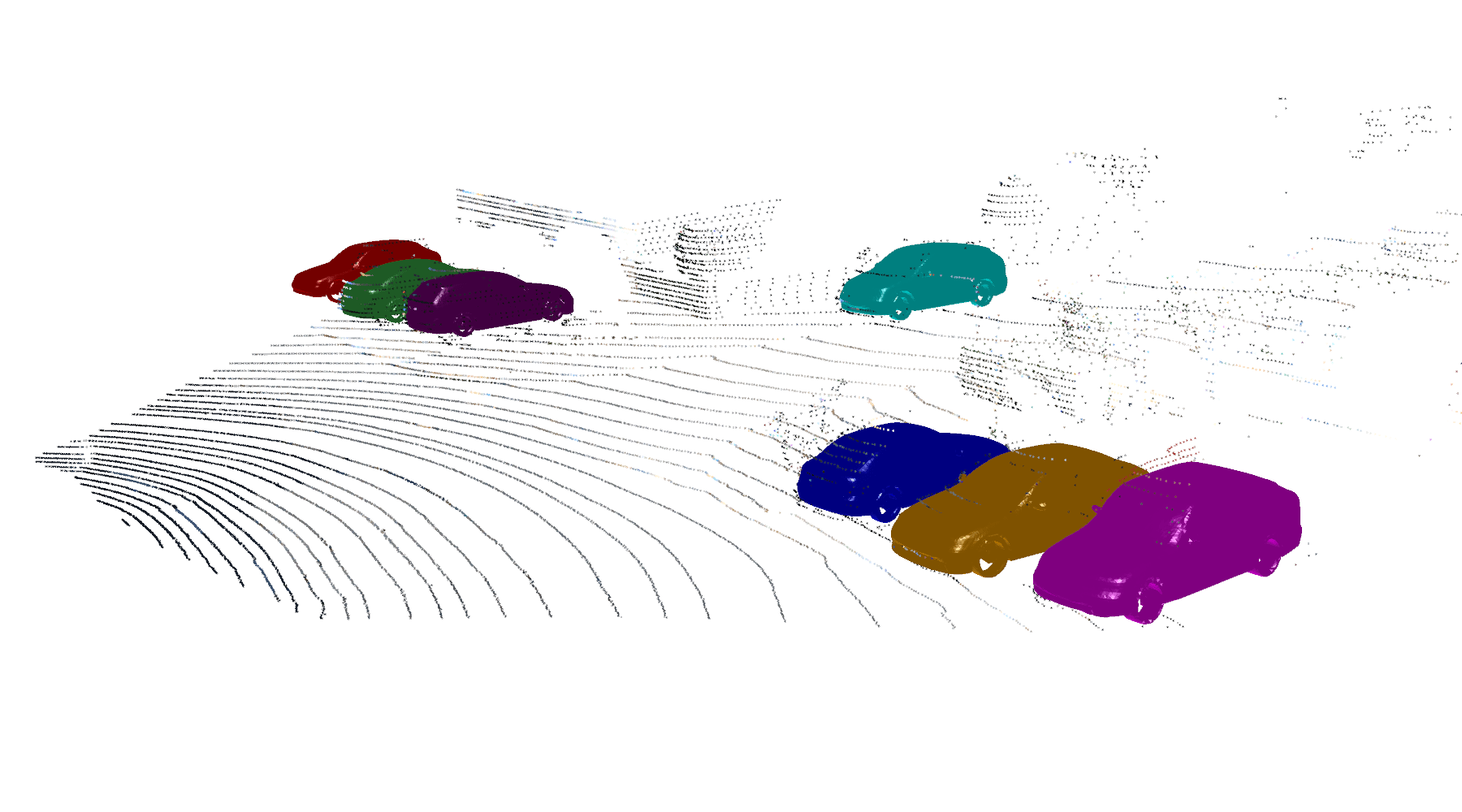}
\includegraphics[width=0.48\textwidth]{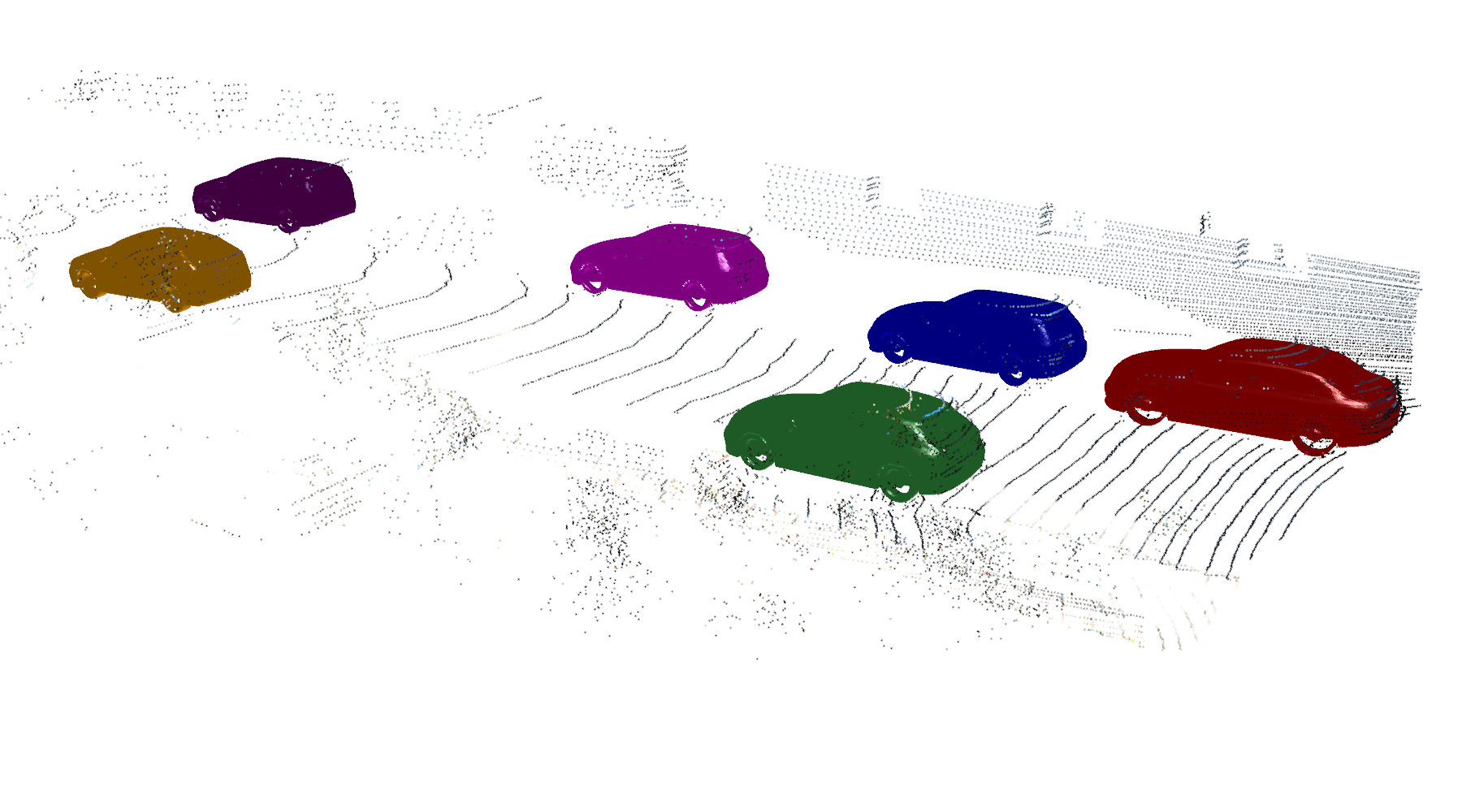}
\caption{More qualitative results on object reconstruction from a single view-point \vspace{-1cm}}
\label{fig:recon_results}
\end{figure*}

\begin{figure*}[tpb]
\centering
\includegraphics[width=0.90\textwidth]{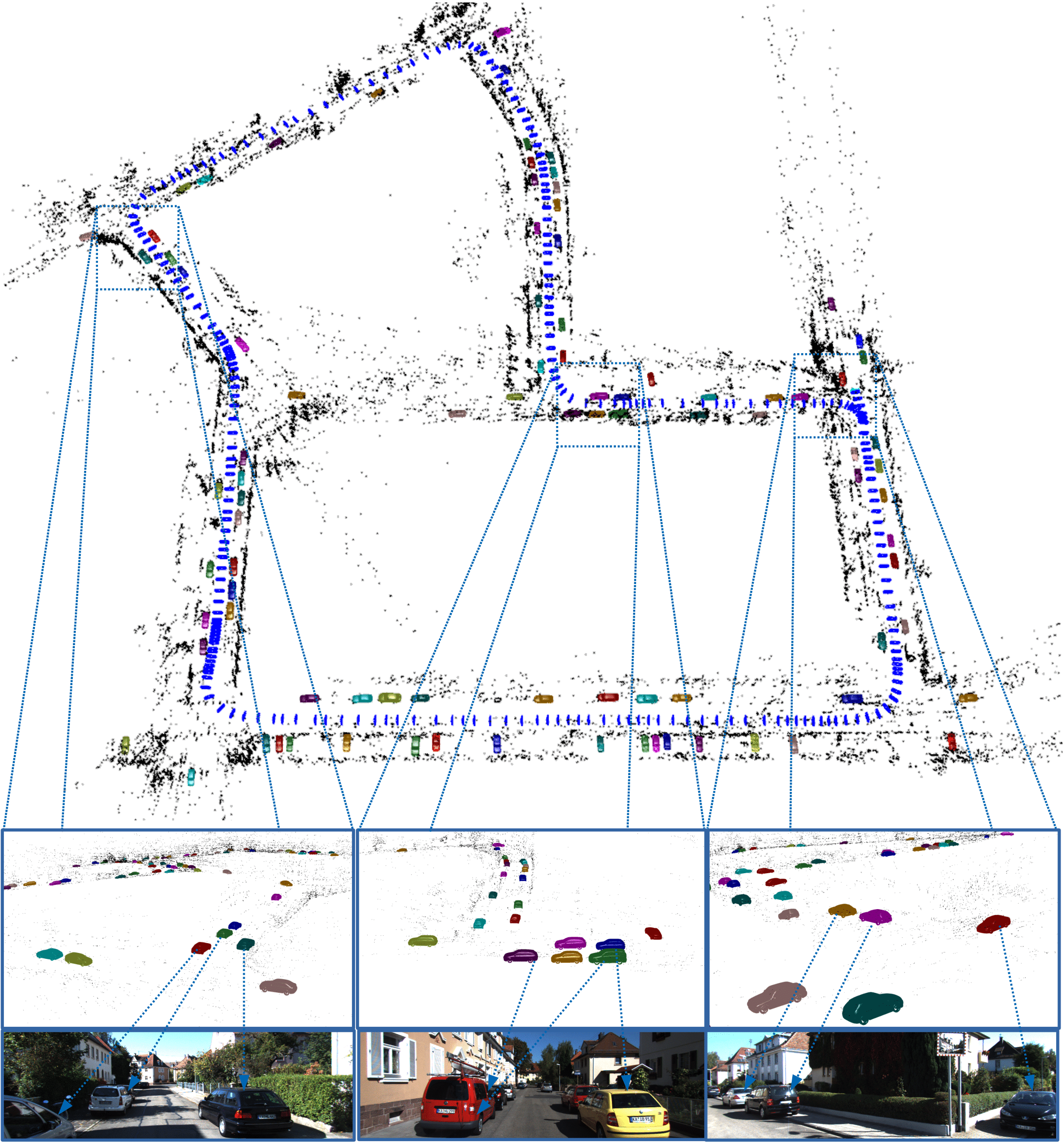}
\caption{Reconstructed map of KITTI-07. Note the dense reconstruction of the objects and the shape variations. \vspace{-1cm}}
\label{fig:map_result}
\end{figure*}

We show more qualitative reconstruction results in Fig.~\ref{fig:recon_results}. For each scene the RGB image is shown on the top and the reconstruction results are shown underneath. We also show an example of reconstructed map in Fig.~\ref{fig:map_result}.

{\small
\bibliographystyle{ieee_fullname}
\bibliography{supp_materials}
}